\documentclass[lettersize,journal]{IEEEtran}
\usepackage{amsmath,amsfonts}
\usepackage{algorithmic}
\usepackage{algorithm}
\usepackage{array}
\usepackage{textcomp}
\usepackage{stfloats}
\usepackage{url}
\usepackage{verbatim}
\usepackage{graphicx}
\usepackage{cite}

\usepackage{url}            
\usepackage{booktabs}       
\usepackage{amsfonts}       
\usepackage{nicefrac}       
\usepackage{microtype}      
\usepackage{xcolor}         
\usepackage{xspace}
\usepackage{graphicx}
\usepackage{amsmath}
\usepackage{amssymb}
\usepackage{enumitem}
\usepackage{amsthm}

\usepackage{subfigure}
\theoremstyle{definition}
\newtheorem{definition}{Definition}[]

\usepackage{color, colortbl}
\definecolor{Gray}{gray}{0.9}

\newcommand{\ourmethod}{{ARC}\xspace}
\newcommand{\ourzero}{{ARC\textsubscript{zero}}\xspace}

\definecolor{fgreen}{RGB}{177,207,149}
\definecolor{fred}{RGB}{234,179,138}

\definecolor{firstcolor}{RGB}{20,128,85}
\definecolor{secondcolor}{RGB}{20,104,168}
\definecolor{thirdcolor}{RGB}{236,84,20}

\newcommand{\first}[1]{\textcolor{firstcolor}{\underline{\mathbf{#1}}}}
\newcommand{\second}[1]{\textcolor{secondcolor}{\underline{#1}}}
\newcommand{\third}[1]{\textcolor{thirdcolor}{\underline{#1}}}

\expandafter\let\csname algorithm*\endcsname\relax
\expandafter\let\csname endalgorithm*\endcsname\relax

\usepackage[linesnumbered,algoruled,boxed,lined,ruled]{algorithm2e}
\SetKwInOut{Param}{Parameters}

\usepackage{pifont}

\definecolor{dgreen}{RGB}{186, 214, 188}
\definecolor{dred}{RGB}{244,200,200}

\usepackage{tcolorbox}

\definecolor{revcolor}{RGB}{0, 0, 0} 
\newcommand{\rev}[1]{\textcolor{revcolor}{#1}}

\begin{document}
\title{From Few-Shot to Zero-Shot: \\Towards Generalist Graph Anomaly Detection}

\author{Yixin Liu, 
Shiyuan Li, 
Yu Zheng,
Qingfeng Chen, 
Chengqi Zhang,~\IEEEmembership{Fellow,~IEEE}, \\ 
Philip S. Yu,~\IEEEmembership{Life Fellow,~IEEE}, Shirui Pan,~\IEEEmembership{Senior Member,~IEEE}%

\IEEEcompsocitemizethanks{
	\IEEEcompsocthanksitem Y. Liu, S. Li, Y. Zheng, and S. Pan are with the School of Information and Communication Technology, Griffith University, Australia.
	\IEEEcompsocthanksitem Q. Chen is with the School of Computer, Electronics and Information, Guangxi University, Nanning, China. 
	\IEEEcompsocthanksitem C. Zhang is with the Department of Data Science and Artificial Intelligence, The Hong Kong Polytechnic University, Hong Kong. 
	\IEEEcompsocthanksitem P. S. Yu is with the Department of Computer Science, University of Illinois at Chicago, Chicago, IL 60607-7053, USA.
	\IEEEcompsocthanksitem Corresponding author: Shirui Pan (E-mail: s.pan@griffith.edu.au).
        \IEEEcompsocthanksitem Y. Liu and S. Li contributed equally to this work.
    }
}

\markboth{Journal of \LaTeX\ Class Files,~Vol.~14, No.~8, August~2021}%
{Shell \MakeLowercase{\textit{et al.}}: A Sample Article Using IEEEtran.cls for IEEE Journals}


\maketitle

\begin{abstract}
Graph anomaly detection (GAD) is critical for identifying abnormal nodes in graph-structured data from diverse domains, including cybersecurity and social networks. The existing GAD methods often focus on the learning paradigms of ``one-model-for-one-dataset'', requiring dataset-specific training for each dataset to achieve optimal performance. However, this paradigm suffers from significant limitations, such as high computational and data costs, limited generalization and transferability to new datasets, and challenges in privacy-sensitive scenarios where access to full datasets or sufficient labels is restricted. \rev{To address these limitations, we propose a novel generalist GAD paradigm that aims to develop a unified model capable of detecting anomalies on multiple unseen datasets without extensive retraining/fine-tuning or dataset-specific customization.} \rev{To this end,} we propose \ourmethod, a few-shot generalist GAD method that leverages \rev{in-context learning and requires only a few labeled normal samples at inference time.} Specifically, \ourmethod consists of three core modules: a feature \underline{A}lignment module \rev{to unify and align features across datasets,} a \underline{R}esidual GNN encoder to capture dataset-agnostic anomaly representations, and a cross-attentive in-\underline{C}ontext learning module \rev{to score anomalies using few-shot normal context.} Building on \ourmethod, we further introduce \ourzero \rev{for the zero-shot generalist GAD setting, which selects representative pseudo-normal nodes via a pseudo-context mechanism and thus enables fully label-free inference on unseen datasets.} Extensive experiments on 17 real-world graph datasets demonstrate that both \ourmethod and \ourzero effectively detect anomalies, exhibit strong generalization ability, and perform efficiently under few-shot and zero-shot settings.

\end{abstract}


\section{Introduction}
\IEEEPARstart{G}{raph} anomaly detection (GAD) aims to detect nodes that deviate significantly from the majority within a graph~\cite{dominant_ding2019deep,cola_liu2021anomaly,bwgnn_tang2022rethinking,qiao2024deep,ma2021comprehensive,miao2025blindguard}. Since graph-structured data are ubiquitous in real-world domains such as finance, cybersecurity, e-commerce, and social networks, GAD holds significant practical value and has received widespread research attention~\cite{caregnn_dou2020enhancing,wang2022wrongdoing}. For example, in the domain of cybersecurity, GAD is commonly used to identify compromised devices or users within a network, such as detecting nodes exhibiting unusual communication patterns that could indicate network intrusions or insider threats~\cite{wu2021graph}. 

\begin{figure}[!t]
  \centering
  \includegraphics[width=0.95\columnwidth]{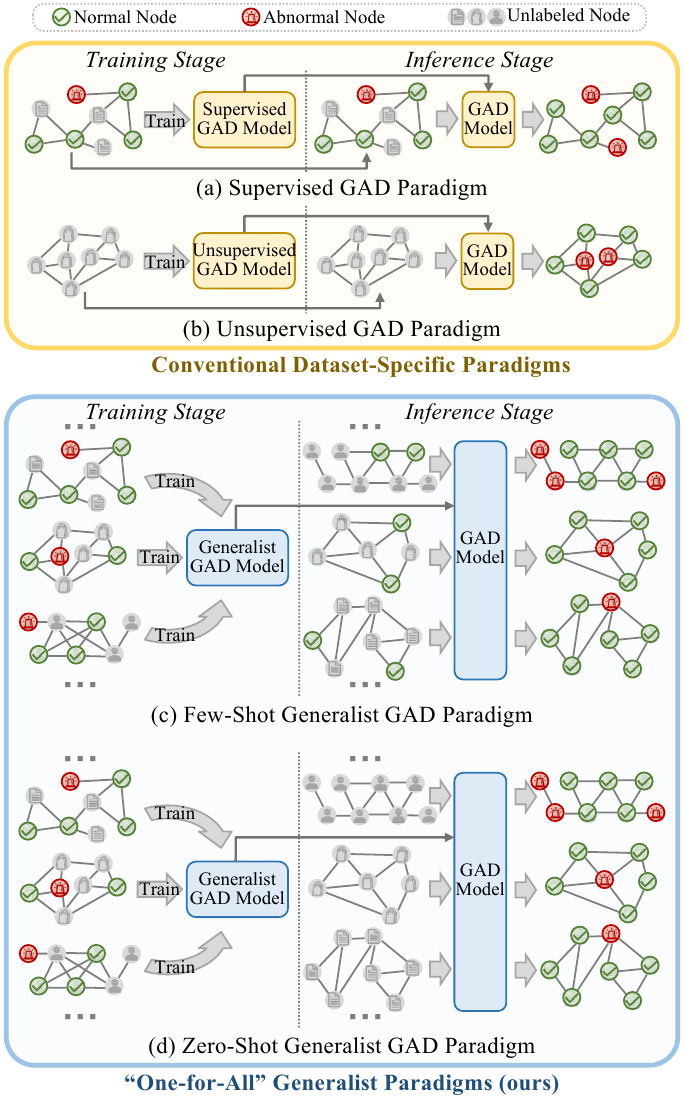}
  \caption{The sketch maps of the existing data-specific GAD paradigms ((a) supervised and (b) unsupervised) and our generalist GAD paradigms ((c) few-shot and (d) zero-shot).}
  \label{fig:sketch}
  \vspace{-3mm}
\end{figure}

With the breakthroughs in deep learning techniques, in recent years, graph neural networks (GNNs)-based GAD methods have emerged as the de facto solution and become the focus of extensive research~\cite{ma2021comprehensive,qiao2024deep,tam_qiao2024truncated}. The general pipeline of these methods involves training a GNN-based GAD model in a data-driven manner, allowing the model to predict an anomaly score for each node that indicates its level of abnormality~\cite{dominant_ding2019deep,cola_liu2021anomaly,bwgnn_tang2022rethinking}. Based on the availability of anomaly labels during training, existing methods can be categorized into two types: supervised GAD methods and unsupervised GAD methods. In supervised methods (Fig.~\ref{fig:sketch}(a)), the GNN-based GAD model is a binary classifier that is trained under the supervision of anomaly labels. With specialized architecture designs and objective functions, these GNN classifiers are highly effective at capturing anomaly patterns~\cite{bwgnn_tang2022rethinking,caregnn_dou2020enhancing,ghrn_gao2023addressing,tang2024gadbench}. Unlike supervised methods, unsupervised GAD methods (Fig.~\ref{fig:sketch}(b)) optimize the models using unsupervised learning techniques without using labels, and then estimate abnormality through carefully designed anomaly score functions~\cite{zhao2025freegad,pan2025survey,pan2026correcting}.

Despite the great success afforded by current GAD methods, they typically follow \textbf{dataset-specific learning paradigms}: the GAD model is trained exclusively on a target dataset and is only effective for that dataset. Unfortunately, the inherent limitations of such ``one-model-for-one-dataset'' paradigms not only increase the training and data costs of GAD methods but also hinder their application to new scenarios. \textit{Firstly}, training a dedicated model from scratch for each target dataset can result in high application costs, including training time and device usage, particularly for large-scale graph data. \textit{Secondly}, training GAD models requires full access to the target dataset, and for supervised methods, sufficient labels are also necessary. This makes GAD models difficult to apply in privacy-sensitive scenarios (e.g., healthcare and finance) where data is usually inaccessible. \textit{Thirdly}, it is non-trivial to select the ``optimal'' GAD method for each dataset, as the performance of GAD models varies across different datasets~\cite{tang2024gadbench}. In such cases, one might need to experiment with various GAD approaches and carefully tuned hyperparameters to adapt to a new dataset, leading to more expensive application costs. 
Given the aforementioned limitations, we raise the core research question of this article: 

\vspace{-0.1cm}
\begin{tcolorbox}
[boxsep=0mm,left=2.5mm,right=2.5mm,colframe=black!55,colback=black!5]
\textbf{Rresearch Question:} \textit{Instead of learning dataset-specific GAD models, can we train a \rev{unified generalist} GAD model that can be applied across multiple datasets without requiring extensive retraining or customization?}
\end{tcolorbox}
\vspace{-0.1cm}

Following the recent research trend of foundation models and artificial general intelligence, a promising answer to this question is to develop a \textbf{generalist GAD method}. As illustrated in the blue block of Fig.~\ref{fig:sketch}, a generalist GAD model is trained on a collection of datasets, learning to capture anomaly patterns across various domains. Once it is well trained, the model can directly detect anomalies in any unseen dataset during the inference phase. The generalist learning paradigm, which has already proven effective in image anomaly detection~\cite{inctrl_zhu2024toward} and link prediction~\cite{unilp_dong2024universal}, naturally avoids the limitations of the traditional approach. On the one hand, the generalist paradigm eliminates the need for full datasets and labels during training, enabling it to efficiently handle large-scale data and privacy-sensitive domains. On the other hand, it reduces the time and costs associated with data-specific model training and complex model selection, making it more adaptable and cost-effective for new scenarios.

According to the availability of labels in the target dataset, the generalist GAD problem can be divided into two types: few-shot and zero-shot settings~\cite{inctrl_zhu2024toward,niu2024zero}. In the \textbf{few-shot} generalist GAD setting (Fig.~\ref{fig:sketch}(c)), the labels of \textit{few-shot normal samples} are available during inference time. These samples provide crucial guidance for the generalist GAD model, offering a reference for normal behaviors and insights into the characteristics of specific datasets. Note that the number of shots can be far fewer than what is typically required for supervised GAD, significantly reducing its application costs. In the \textbf{zero-shot} setting (Fig.~\ref{fig:sketch}(d)), differently, the generalist GAD model can detect anomalies without needing any labeled samples from the target dataset during inference. For the scenarios where annotation costs are extremely high, a zero-shot generalist GAD model can serve as a complementary approach to its few-shot counterpart, further expanding its applicability in real-world scenarios. Since both few-shot and zero-shot settings are commonly encountered in real-world applications, the goal of this paper is to develop effective solutions tailored to both of them.

While generalist GAD paradigms have strong applicability in real-world scenarios, designing an effective generalist GAD method remains a non-trivial task. Specifically, four key challenges arise, as outlined below:

\begin{itemize}[leftmargin=*, itemsep=0pt, parsep=0pt, topsep=2pt]
    \item \textit{\textbf{Challenge 1} - feature diversity across different datasets}. Unlike image or language data where raw features are mapped into a unified space, in graph-structured data, the dimensionality and semantic space of node attributes can differ significantly across diverse datasets. In this case, aligning the features of different datasets at the input side of the generalist GAD model becomes a crucial challenge~\cite{gcope_zhao2024all}. 
    \item \textit{\textbf{Challenge 2} - anomaly-aware representation encoding}. In a GAD model, the quality of anomaly-aware node-level representations is largely depends on the specific design of GNN architectures. In generalist settings, how to improve the GNN encoder to capture dataset-agnostic representations presents a significant challenge, as anomaly patterns can vary greatly across different datasets.
    \item \textit{\textbf{Challenge 3} - few-shot normal sample-guided detection}. Under the conventional paradigms, existing GAD methods can effectively learn how to identify anomalies through dataset-specific training. However, a few-shot generalist GAD model is expected to detect anomalies with minimal labeled samples and without model optimization, necessitating the design of models that can effectively leverage the limited few-shot normal samples. 
    \item \textit{\textbf{Challenge 4} - pattern identification in zero-shot scenarios}. In the zero-shot setting, the generalist GAD model cannot rely on any labeled samples to obtain information about the data characteristics and distribution of the target dataset. Under this circumstance, accurately recognizing and capturing these characteristics becomes a significant challenge in designing an effective zero-shot method.
    
\end{itemize}

To tackle these challenges, in this work\footnote{It is an extension and re-innovation to our conference article~\cite{liu2024arc} accepted by NeurIPS 2024.}, we \rev{take a pioneering step in studying} the generalist GAD problem. We propose \ourmethod and \ourzero, two effective generalist GAD methods tailored for the few-shot and zero-shot paradigms, respectively. Specifically, we first present the few-shot method \ourmethod which is composed of three well-crafted modules. To solve \textit{\textbf{Challenge~1}}, we design a smoothness-based feature \textbf{\underline{A}}lignment module. The alignment module first unifies the feature dimensions of different datasets using linear projections, and then aligns their semantic space by employing a smoothness-based reordering technique to ensure consistency across datasets. To address \textit{\textbf{Challenge~2}}, we propose an ego-neighbor \textbf{\underline{R}}esidual GNN as our encoder. Employing a specialized multi-hop residual architecture, the encoder learns anomaly-aware representations that capture both high-order affinity and heterophily in diverse datasets. To deal with \textit{\textbf{Challenge~3}}, we design a cross-attentive in-\textbf{\underline{C}}ontext anomaly scoring module. A cross-attention block is established to reconstruct the representations of unlabeled nodes using the representations of few-shot normal nodes (denoted as context nodes). The reconstruction error for each node is then used as its anomaly score, allowing \ourmethod to make predictions in an in-context learning manner. Last but not least, to address \textit{\textbf{Challenge~4}}, we propose \ourzero, a zero-shot generalist GAD method that builds upon and extends the framework of \ourmethod. In \ourzero, we introduce a pseudo-context mechanism that automatically and iteratively selects the most representative samples to act as context nodes in \ourmethod. This self-learning approach allows the model to capture dataset characteristics without any labeled data, enabling effective zero-shot anomaly detection. In summary, the primary contributions of this work are as follows:

\begin{itemize}[leftmargin=*, itemsep=0pt, parsep=0pt, topsep=2pt]
    \item \textbf{New research problem.} \rev{We formulate and investigate the generalist GAD problem, which is an important yet under-explored direction.} We further introduce two generalist GAD paradigms (few-shot and zero-shot), which model different real-world scenarios with varying levels of label availability.
    \item \textbf{Few-shot solution.} Aiming at the few-shot generalist GAD problem, we propose a novel method \ourmethod. With carefully designed alignment, encoding, and in-context learning modules, \ourmethod can efficiently detect anomalies in unseen graphs on the fly using only a few shots of normal samples.
    \item \textbf{Zero-shot solution.} For the more challenging zero-shot problem, we further propose \ourzero, which enhances \ourmethod by incorporating a pseudo-context mechanism. \ourzero can automatically identify representative samples as context nodes, enabling zero-shot detection in new datasets. 
    \item \textbf{Extensive experiments.} We conduct extensive experiments on 17 real-world graph datasets (4 for training and 13 for evaluation), and the experimental results indicate the anomaly detection performance, out-of-domain generalization ability, running efficiency, and few-shot/zero-shot capability of \ourmethod and \ourzero. 
\end{itemize}

This work significantly extends our preliminary paper published in NeurIPS conference~\cite{liu2024arc} by providing sufficient improvements in several key aspects: 1) We formulate the generalist GAD problem into a finer-grained taxonomy, categorizing it into few-shot and zero-shot settings. These new settings align with real-world constraints on label availability and application scenarios, making them more practical for diverse anomaly detection tasks. 2) We propose a novel method \ourzero for zero-shot generalist GAD problem. Extensive experiments verify the superior zero-shot performance of \ourmethod. 3) We provide a more in-depth analysis and discussion of the design motivation and underlying mechanisms of \ourmethod and \ourzero. 4) We conduct more experiments to discuss the performance and characteristics of the proposed methods, including out-of-domain generalization experiments, detailed ablation studies, parameter analysis, and visualization.

\section{Related Works}

\subsection{Anomaly Detection}
{Anomaly detection (AD) aims to identify instances that deviate significantly from the norm~\cite{ahmed2021graph}. Due to limited anomaly labels, most AD methods operate in unsupervised settings~\cite{ruff2018deep,xiang2024exploiting}, including one-class modeling~\cite{ruff2018deep}, reconstruction-based approaches~\cite{zhou2017anomaly}, and distance/generative/self-supervised techniques~\cite{roth2022towards,defard2021padim,sehwag2021ssd}. Recently, large language models (LLMs) and visual-language models (VLMs) also show potential in addressing AD tasks~\cite{jeong2023winclip,su2024large}.}

{However, most AD methods are dataset-specific, limiting generalization to unseen domains. 
Generalist AD targets a single model that transfers to new datasets without retraining~\cite{zhou2023anomalyclip,inctrl_zhu2024toward}, e.g., InCTRL using pre-trained VLMs~\cite{inctrl_zhu2024toward}.
While the above methods focus on generalist AD for image data, in this paper, we propose the first generalist AD method for graph-structured data.} 
%


\subsection{Graph Anomaly Detection}

{Graph anomaly detection (GAD) aims to identify entities that deviate from the majority in graphs~\cite{bwgnn_tang2022rethinking,tang2024gadbench}. In this paper, we focus on node-level detection~\cite{ma2021comprehensive}. While early studies used shallow mechanisms (e.g., residual analysis)~\cite{li2017radar,peng2018anomalous}, recent approaches predominantly rely on graph neural networks (GNNs)~\cite{bwgnn_tang2022rethinking,qiao2024deep}, which can be broadly categorized into supervised and unsupervised GAD methods~\cite{ma2021comprehensive,tang2024gadbench}.} 
{Supervised GAD formulates detection as binary classification and improves GNN aggregation/architecture/losses for anomaly patterns~\cite{bwgnn_tang2022rethinking,caregnn_dou2020enhancing,he2021bernnet,ghrn_gao2023addressing}, e.g., BWGNN with band-pass filters and GHRN with high-frequency pruning~\cite{bwgnn_tang2022rethinking,ghrn_gao2023addressing}.}
{Since labels are often unavailable, unsupervised GAD methods detect anomalies without annotated labels~\cite{fan2020anomalydae,cola_liu2021anomaly,huang2022hop,pan2023prem}. 
Representative approaches include reconstruction-based methods such as DOMINANT~\cite{dominant_ding2019deep}, contrastive learning methods such as CoLA~\cite{cola_liu2021anomaly}, and affinity-based methods such as TAM~\cite{tam_qiao2024truncated}.}

\rev{Although the methods mentioned above can detect anomalies in the training dataset, they are typically ``one-for-one'': a separate model (and often dataset-specific training/validation) is required for each target graph, which limits their applicability to new and unseen domains. In contrast, cross-domain and transfer-learning GAD aim to improve generalization across domains by leveraging a source domain and adapting to a specific target domain, where target-domain adaptation or fine-tuning is commonly involved~\cite{ding2021cross,wang2023cross}. However, these settings still assume access to the target domain and do not directly support on-the-fly deployment to arbitrary unseen graphs.}

\rev{Motivated by this gap, we study generalist GAD, which trains a single model on a collection of graphs and directly deploys it to unseen graphs without any dataset-specific fine-tuning. Very recently, AnomalyGFM~\cite{qiao2025anomalygfm} and UNPrompt~\cite{niu2024zero} explore generalist GAD by leveraging GAD-oriented foundation modeling and unified neighborhood prompts, respectively. Nonetheless, their performance can be sensitive to design choices and hyperparameter settings across datasets, which may hinder truly prior-free deployment to new domains. To address this, \ourmethod~\cite{liu2024arc} proposes an attention-based in-context learning mechanism for few-shot generalist GAD with a small set of normal nodes at inference time, and we further develop \ourzero to remove this requirement by constructing a pseudo-context and refining it iteratively with multi-round aggregation for reliable zero-shot anomaly scoring.}

\subsection{In-Context Learning}
{In-context learning (ICL) adapts a model at inference time using context examples, without explicit retraining~\cite{dong2022survey}. It has shown effectiveness in natural language processing and vision, and is widely used in few-shot/zero-shot settings~\cite{alayrac2022flamingo,bar2022visual}.}
{Recent work extends ICL to graphs, e.g., prompt graphs for node/edge tasks~\cite{huang2024prodigy} and universal link prediction with contextual cues~\cite{unilp_dong2024universal}. However, they typically assume a multi-class context at inference, whereas generalist GAD needs a one-class (normal) context; our attention-based ICL mechanism addresses this and further supports zero-shot detection via pseudo-context refinement.}

\section{Background and Problem Statement}

\subsection{Notations}

Throughout this paper, we use bold lowercase letters (e.g.~$\mathbf{a}$), bold uppercase letters (e.g.~$\mathbf{A}$), and calligraphic fonts (e.g.~$\mathcal{A}$) to denote vectors, matrices, and sets, respectively. 

We denote an undirected and attributed graph as $G=(\mathcal{V}, \mathcal{E}, \mathbf{X})$, where $\mathcal{V} = \{v_1, \cdots, v_n\}$ is the node set with $n$ nodes, $\mathcal{E}$ is the edge set with $m$ edges, and $\mathbf{X} \in \mathbb{R}^{n \times d}$ is the feature matrix where the $i$-th row $\mathbf{x}_i$ is the feature vector for node $v_i$. The graph structure can be represented by a binary adjacency matrix $\mathbf{A} \in \{0,1\}^{n \times n}$, where the $i,j$-th entry $a_{i,j}=1$ indicates that $v_i$ and $v_j$, otherwise $a_{i,j}=0$. 

In the context of anomaly detection (GAD) tasks, the node set can be written by $\mathcal{V}=\mathcal{V}_a \cup \mathcal{V}_n$ (satisfying $\mathcal{V}_a \cap \mathcal{V}_n=\emptyset$), where $\mathcal{V}_a$ and $\mathcal{V}_n$ are the abnormal node set and normal node set, respectively. Typically, the number of anomalies is far less than that of normal nodes, i.e., $|\mathcal{V}_a| \ll |\mathcal{V}_n|$. A label vector $\mathbf{y} \in \{0,1\}^n$ can be used to describe node abnormality, where the $i$-th element $y_i=1$ \textit{iff} $v_i \in \mathcal{V}_a$ and $\mathbf{y}_i=0$ \textit{iff} $v_i \in \mathcal{V}_n$. 

\subsection{Problem definition}


\theoremstyle{definition}
\begin{definition}[Dataset-Specific GAD Problem]
\label{def:gad}
Given graph $G=(\mathcal{V},\mathcal{E},\mathbf{X})$, the objective of dataset-specific GAD is to learn an anomaly scoring function (i.e., GAD model) $f: \mathcal{V} \rightarrow \mathbb{R}$ such that $f(v^{\prime}) > f(v)$ for $\forall v^{\prime} \in \mathcal{V}_a$ and $\forall v \in \mathcal{V}_n$. The GAD model $f$ is optimized on the dataset $\mathcal{D}=({G},\mathbf{y})$ during the training phase and is then used solely to predict the abnormality of nodes in the same dataset $\mathcal{D}$ during the inference phase. For supervised GAD methods, the labels of a fraction of nodes are available for training, while unsupervised GAD methods solely learn from the unlabeled graph ${G}$.
\end{definition}

The conventional data-specific setting follows the paradigm of ``one-model-for-one-dataset'', which comes with several inherent limitations such as high training costs, excessive data requirements, and poor generalization ability. To address these limitations, this paper investigates the generalist GAD problem, aiming to develop a \rev{unified generalist} model that can be directly applied to multiple and diverse datasets. Specifically, the generalist GAD problem can be written as:

\theoremstyle{definition}
\begin{definition}[Generalist GAD Problem]
\label{def:gen_gad}
We define $\mathcal{T}_{train}=\{\mathcal{D}^{(1)}_{train}, \cdots, \mathcal{D}^{(N)}_{train}\}$ as a collection of training datasets, where each $\mathcal{D}^{(i)}_{train}=(\mathcal{G}^{(i)}_{train},\mathbf{y}^{(i)}_{train})$ is a labeled dataset from an arbitrary domain. Similarly, a collection of testing datasets can be defined as $\mathcal{T}_{test}=\{\mathcal{D}^{(1)}_{test}, \cdots, \mathcal{D}^{(N')}_{test}\}$, where each graph may come from a different domain. Note that there is no overlap between two collections, i.e. $\mathcal{T}_{train} \cap \mathcal{T}_{test}=\emptyset$. The objective of generalist GAD is to train a generalist GAD model $f$ on $\mathcal{T}_{train}$, so that $f$ can identify anomalies in any testing dataset $\mathcal{D}^{(i)}_{test} \in \mathcal{T}_{test}$, without the need for dataset-specific re-training or fine-tuning.
\end{definition}

According to the availability of the labels of testing datasets during inference, the generalist GAD problem can be further divided into two categories: \textbf{few-shot} generalist GAD and \textbf{zero-shot} generalist GAD. In the few-shot setting, a small number of normal node labels from each dataset $\mathcal{D}^{(i)}_{test}$ are available during inference~\cite{inctrl_zhu2024toward}. In the zero-shot setting, the labels of the testing dataset are completely inaccessible to the GAD model, making it more challenging than its few-shot counterpart~\cite{niu2024zero}. In this paper, we propose simple yet effective solutions — \ourmethod and \ourzero — to address both settings, respectively.

\section{ARC: A Few-Shot Generalist GAD Approach}
\begin{figure*}[!t]
  \centering
  \includegraphics[width=\textwidth]{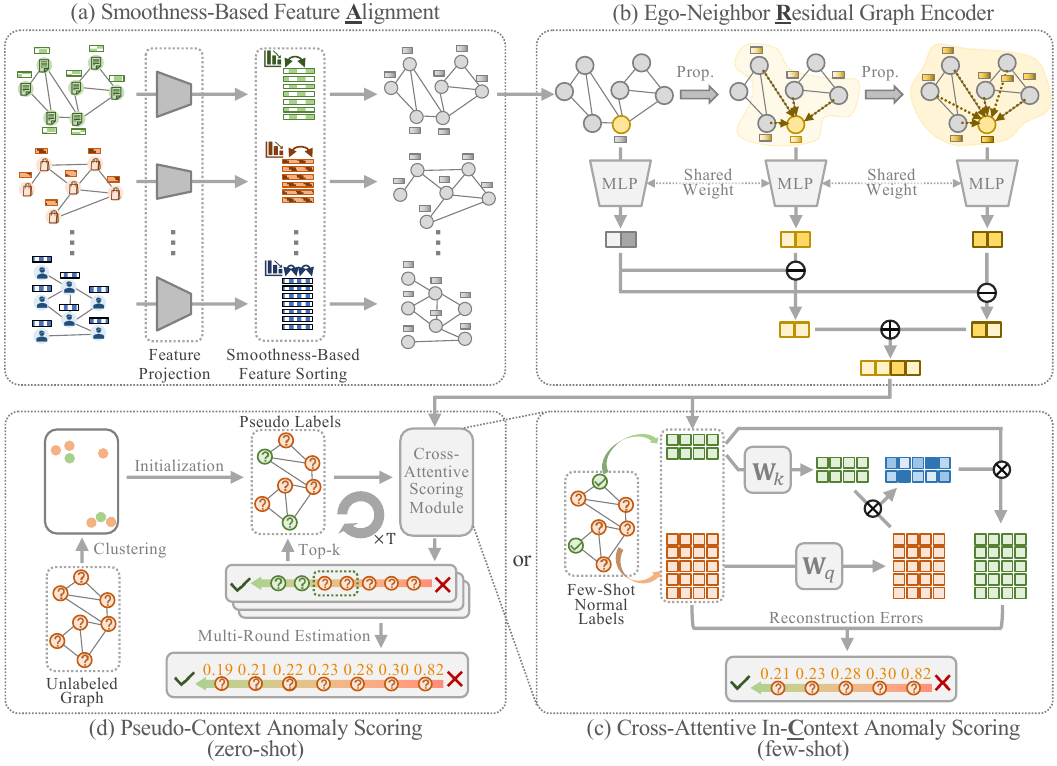}
  \caption{The overall pipeline of the proposed methods \ourmethod, illustrating by (a)\ding{220}(b)\ding{220}(c), and \ourzero, illustrating by (a)\ding{220}(b)\ding{220}(d).}
  \label{fig:pipeline}
  \vspace{-3mm}
\end{figure*}

In this section, we introduce ARC, a few-shot generalist GAD approach. As illustrated by the pipeline (a)\ding{220}(b)\ding{220}(c) in Fig.~\ref{fig:pipeline}, \ourmethod is composed of three core modules: smoothness-based feature alignment, ego-neighbor residual graph encoder, and cross-attentive in-context anomaly scoring. In the initial step, we align the features from various datasets through projection and sorting, ensuring consistency in both dimensionality and semantics. Subsequently, a residual GNN with multi-hop information aggregation is employed to capture the representation of each node. Finally, a cross-attentive anomaly scoring module reconstructs the representations of query nodes (i.e., unlabeled nodes) by leveraging the representations of context nodes (i.e., few-shot normal nodes) through a cross-attention block, where the reconstruction errors finally serve as the predicted anomaly scores. In the following subsections, we provide a detailed introduction to the algorithmic design of each module in \ourmethod. 

\subsection{Smoothness-Based Feature Alignment}

A primary requirement for establishing a generalist and universal model for multiple datasets is that they share a unified input space. For example, in computer vision tasks, images are typically represented as pixel grids with consistent dimensionality, and in natural language processing, texts are tokenized into a common vocabulary space. Unlike these scenarios, in graph-structured data, the vectorized node features from different datasets often exhibit varying dimensions and belong to distinct semantic spaces. For instance, a social network dataset might have nodes with social attributes like age and profession, while an e-commerce network could feature nodes with attributes such as purchase frequency and average spending. To alleviate this feature heterogeneity, the first step in our generalist GAD method is to align the features of different datasets by standardizing their dimensionality and semantic space. Specifically, the dimensionality is unified using a linear feature projection, while the semantic space alignment is achieved through smoothness-based feature sorting, which will be demonstrated as follows.

\subsubsection{Linear Feature Projection}
To align the dimensionality of features from different graph datasets, we employ a dataset-specific linear feature projection for each of them. Concretely, given a feature matrix $\mathbf{X}^{(i)} \in \mathbb{R}^{n^{(i)} \times d^{(i)}}$ from $\mathcal{D}^{(i)} \in \mathcal{T}_{train} \cup \mathcal{T}_{test}$, a linear mapping is applied to project the feature into a unified dimension $d_u$:

\begin{equation}
\label{eq:projection}
\tilde{\mathbf{X}}^{(i)} = \operatorname{Proj}\left(\mathbf{X}^{(i)}\right) = \mathbf{X}^{(i)} \mathbf{W}^{(i)},  
\end{equation}

\noindent where $\tilde{\mathbf{X}}^{(i)} \in \mathbb{R}^{n^{(i)} \times d_u}$ is the projected feature matrix for $\mathcal{D}^{(i)}$ and $\mathbf{W}^{(i)} \in \mathbb{R}^{d^{(i)} \times d_u}$ is a dataset-specific linear projection weight matrix. 

A straightforward strategy is to utilize learnable linear mappings to optimize the projection weight matrix for each dataset. However, this approach introduces additional fine-tuning costs for testing datasets, which contradicts the core principle of generalist GAD paradigms. Therefore, to avoid test-time training while maintaining model generality, we employ principal component analysis (PCA)~\cite{pca_abdi2010principal}, a widely used dimensionality reduction method, to conduct the linear mapping. Projecting the features with PCA not only reduces the application costs on testing datasets but also ensures the most informative components of the features across different datasets are captured.

\subsubsection{Smoothness-based Feature Sorting}

After unifying the dimensionality, the next step is to align the feature semantic space across different datasets. To achieve this, one potential approach is to leverage pre-trained language models or large language models (LLMs) to map the features into a common semantic space~\cite{ofa_liu2023one}. Nevertheless, the language model-based strategies require the node features to be naturally represented as texts, which may restrict their applicability to certain types of datasets, especially when node features are non-textual or highly structured, such as numerical or categorical data~\cite{gcope_zhao2024all}. 

Considering the difficulty of semantic-level feature matching, in \ourmethod, we explore an alternative pathway: a\textit{ligning features based on their contribution to the anomaly detection task}. Concretely, we measure the intrinsic importance of specific features for anomaly detection and reorder the permutation of features based on the measured significance. In this way, the features from different datasets can be semantically aligned in a task-sensitive manner, ensuring that the downstream encoder and scoring modules can distinguish and effectively process features with varying contributions to anomaly detection. 

\noindent\textbf{Smoothness as Contribution Measurement.} To evaluate the feature contribution to GAD, a simple and dataset-agnostic measurement is desired. Motivated by its correlations to graph spectral and spatial properties~\cite{bwgnn_tang2022rethinking,ghrn_gao2023addressing}, we propose \textbf{feature-level smoothness} as a key criterion for evaluating feature importance. Formally, given $\mathcal{G}=(\mathcal{V},\mathcal{E},\mathbf{X})$ with a normalized feature matrix $\mathbf{X}$, the smoothness of the $k$-th feature dimension can be formulated by:

\begin{equation}
\label{eq:smoothness}
s_k(\mathbf{X}) = - \frac{1}{|\mathcal{E}|} \sum_{(v_i,v_j) \in \mathcal{E}} \left( \mathbf{X}_{ik} - \mathbf{X}_{jk} \right)^2,
\end{equation}

\noindent where a smaller value of $s_k$ indicates a more significant change in the $k$-th feature between connected nodes in a graph.

Theoretically, we can understand the significance of smoothness to GAD tasks from both spectral and spatial perspectives. 
From a spectral perspective, the inverse of the low-frequency energy ratio is positively correlated with the node-level anomaly degree, as suggested in~\cite{bwgnn_tang2022rethinking}. This means that a high-frequency graph signal (i.e., graph feature) can contribute more to indicating the presence of anomalies. \rev{Mathematically, let $\mathbf{x}_k=\mathbf{X}_{:k}$ and $\mathbf{L}=\mathbf{D}-\mathbf{A}$ be the combinatorial graph Laplacian. Since $\mathbf{x}_k^\top \mathbf{L}\mathbf{x}_k=\frac{1}{2}\sum_{(v_i,v_j)\in\mathcal{E}}(\mathbf{X}_{ik}-\mathbf{X}_{jk})^2$, our smoothness metric is equivalent to the negative normalized Dirichlet energy: $s_k(\mathbf{X})=-\frac{2}{|\mathcal{E}|}\mathbf{x}_k^\top \mathbf{L}\mathbf{x}_k$. Using the spectral decomposition $\mathbf{L}=\mathbf{U}\boldsymbol{\Lambda}\mathbf{U}^\top$, we have $\mathbf{x}_k^\top \mathbf{L}\mathbf{x}_k=\sum_r \lambda_r\langle \mathbf{u}_r,\mathbf{x}_k\rangle^2$. This implies that a lower smoothness score $s_k$ corresponds to higher energy in high-frequency spectral components~\cite{dong2021adagnn}, which are typically indicative of local anomalies.} 
From a spatial perspective, smoothness reflects the local homophily and heterophily of features among neighboring nodes, as the definitions of homophily rate and smoothness are similar~\cite{zheng2022graph}. Since heterophily properties are highly indicative in discriminating anomalies~\cite{ghrn_gao2023addressing,tam_qiao2024truncated}, smoothness serves as a useful metric for evaluating the discriminative power of features in GAD tasks.

\begin{figure*}[t!]
\centering
 \subfigure[Cora]{
   \includegraphics[height=0.212\textwidth]{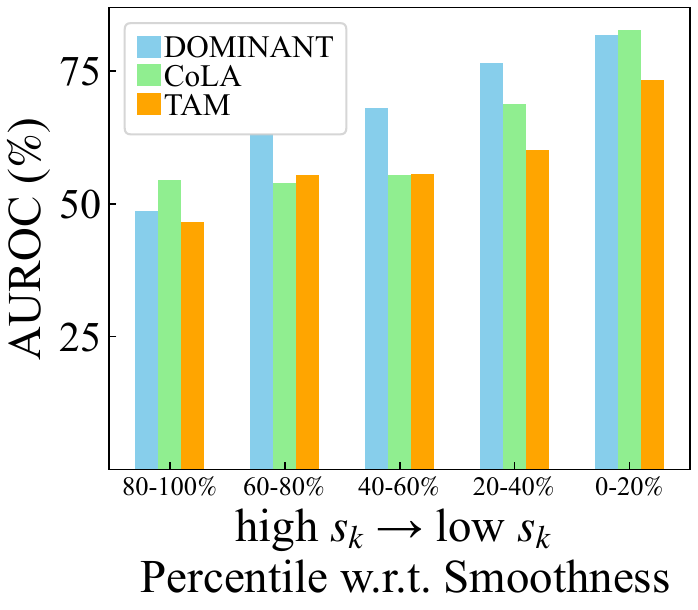}
}
 \hfill
 \subfigure[ACM]{
   \includegraphics[height=0.212\textwidth]{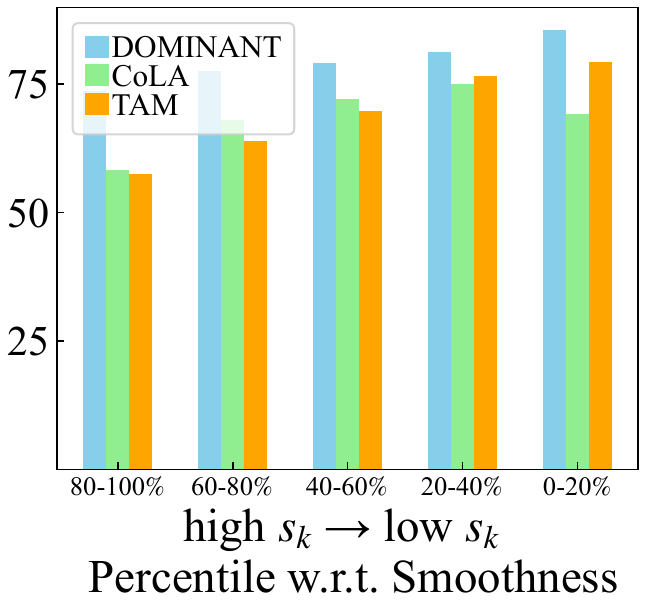}
}
\hfill
 \subfigure[BlogCatalog]{
   \includegraphics[height=0.212\textwidth]{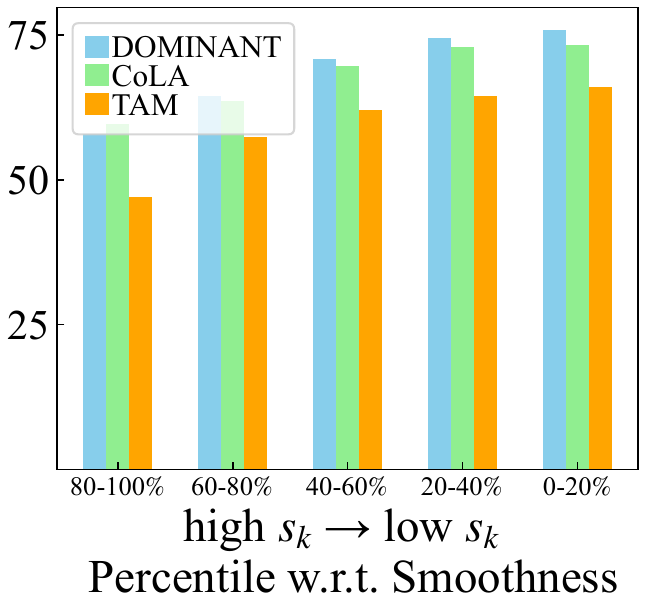}
}
 \hfill
 \subfigure[Facebook]{
   \includegraphics[height=0.212\textwidth]{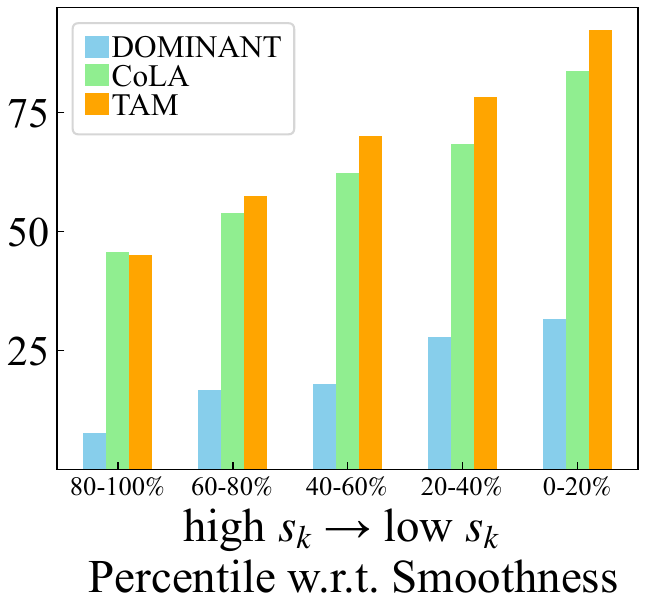}
}
\caption{AUROC on data with 5 groups of feature subsets.}
\label{fig:moti}
\end{figure*}

Empirically, we design a motivating experiment to validate the correlation between a feature's smoothness and its contribution to GAD tasks. During the data preprocessing phase, we calculate $s_k$ of each raw feature dimension and sort the features in descending order. Based on this sorted order, we partition the features into five groups of subsets, corresponding to the percentiles 80\%-100\%, 60\%-80\%, \(\cdots\), and 0\%-20\%. We evaluate the anomaly detection performance of three mainstream GAD methods (DOMINANT~\cite{dominant_ding2019deep}, CoLA~\cite{cola_liu2021anomaly}, and TAM~\cite{tam_qiao2024truncated}) by using each subset of features as input, and the results are shown in Fig.~\ref{fig:moti}. On all four datasets, a consistent trend emerges: feature subsets with lower $s_k$ demonstrate greater effectiveness in discriminating anomalies. In light of this, smoothness has been shown to be an effective anomaly-related indicator for evaluating the relevance of each feature in different datasets.

\noindent\textbf{Smoothness-based Sorting.} 
In \ourmethod, we utilize feature-level smoothness as a metric to reorder the projected features, facilitating task-oriented semantic alignment across datasets. Given the projected feature matrix $\tilde{\mathbf{X}}^{(i)}$, we first calculate the smoothness $s_k(\tilde{\mathbf{X}}^{(i)})$ of each column and then reorder the columns in descending order of their smoothness. The reordered feature matrix, denoted by $\mathbf{X}'^{(i)}$, serves as the aligned feature matrix for the graph dataset $\mathcal{D}^{(i)}$ and is fed into the encoder. 

With the alignment strategy, we ensure that the first-column feature of all datasets is the one with the lowest smoothness, which may have the greatest contribution to the GAD predictions; in contrast, features with higher smoothness will appear in later columns, as they are less critical for anomaly detection. Taking the aligned features as input, the downstream encoder and scoring modules can learn to prioritize and effectively process the most relevant features for anomaly detection. Once a testing dataset is aligned in the same manner, the knowledge and capability of anomaly detection can be seamlessly transferred to this unseen dataset.

\subsection{Ego-Neighbor Residual Graph Encoder}

As the core component of a GAD method, the GNN-based encoder plays a pivotal role in learning representation for each node, capturing structural and semantic information of nodes to support effective anomaly detection. While conventional GNN models (e.g., GCN~\cite{gcn_kipf2017semi} and GAT~\cite{gat_velivckovic2018graph}) perform well on tasks like classification and clustering, they do not inherently capture the nuanced patterns (e.g., high-frequency signals and heterophily) required for anomaly detection, often leading to sub-optimal performance~\cite{tang2024gadbench}. While recent studies have developed specialized GNN architectures for supervised GAD, they are designed for dataset-specific scenarios rather than generalist GAD problems. For generalist scenarios, how to construct a GNN encoder that can capture transferable knowledge across domains remains an open problem.

In this paper, we propose an ego-neighbor residual graph encoder (RGE for short) to learn representations in generalist GAD scenarios. Combining a multi-hop mixing architecture and a residual operation, RGE is able to capture transferable and anomaly-sensitive knowledge on different datasets. As shown in Fig.~\ref{fig:pipeline}(b), RGE employs a ``propagation-then-transformation'' design, which is similar to SGC~\cite{sgc_wu2019simplifying}. Given the input feature matrix $\mathbf{X}^{[0]}$ (here $\mathbf{X}^{[0]}$ is the aligned feature $\mathbf{X}'$ for each dataset), RGE first propagates the node features along the graph structure for $L$ iterations, and then applies a shared MLP to perform non-linear transformations on both the initial and propagated features: 

\begin{equation}
\label{eq:prop_trans}
\mathbf{X}^{[l]} = \tilde{\mathbf{A}}\mathbf{X}^{[l-1]}, \quad \mathbf{Z}^{[l]} = \operatorname{MLP}\left(\mathbf{X}^{[l]}\right), 
\end{equation}

\noindent where $l \in \{0, \cdots, L\}$ is the iteration index, $\mathbf{X}^{[l]}$ is the propagated feature matrix at the $l$-th iteration, $\mathbf{Z}^{[l]}$ is the transformed representation matrix at the $l$-th iteration, and $\tilde{\mathbf{A}}$ is the normalized adjacency matrix~\cite{gcn_kipf2017semi,sgc_wu2019simplifying}. To fully leverage the information at different receptive fields, RGE learns representations from the initial features and all intermediate propagated features, which allows \ourmethod to identify varied anomaly patterns from different datasets. Moreover, we utilize a shared MLP for transformation at all iterations, ensuring that the propagated features are consistently mapped into a unified representation space and preserving low computational costs.

After computing the transformed representations at different propagation steps, i.e., ${\mathbf{Z}^{[0]}, \cdots,\mathbf{Z}^{[L]}}$, residual operations are applied to model the ego-neighbor differences. To be more specific, for each transformed representation matrix $\mathbf{Z}^{[l]}$ ($1 \leq l \leq L$), a residual representation matrix is obtained by calculating the difference between $\mathbf{Z}^{[l]}$ and the ego transformed representation matrix $\mathbf{Z}^{[0]}$ as follows:  

\begin{equation}
\label{eq:residual}
\mathbf{R}^{[l]} = \mathbf{Z}^{[l]} - \mathbf{Z}^{[0]} \quad l \in \{1, \cdots, L\}, 
\end{equation}

\noindent where $\mathbf{R}^{[l]}$ is the residual matrix at the $l$-th propagated iteration. The residual operation is the key design of RGE, allowing \ourmethod to effectively learn representations that are highly sensitive to anomalies. \textit{Firstly}, the residual operation emphasizes the differences between each node and its neighbors, explicitly modeling local affinity through the learned embeddings~\cite{tam_qiao2024truncated}. Since local affinity can effectively reflect node-level abnormalities~\cite{tam_qiao2024truncated,cola_liu2021anomaly}, the residual representations can capture these distinctive features for the downstream anomaly scoring module. \textit{Secondly}, the first-order residual operation ($\mathbf{R}^{[1]}$) can be seen as a high-pass filter on the graph data, functioning similarly to a graph convolution with the graph Laplacian, a well-known high-pass operator on graphs~\cite{dong2021adagnn,gcn_kipf2017semi}. High-pass filters are effective in capturing high-frequency signals and local heterophily, both of which are valuable for detecting anomalies in GAD~\cite{ghrn_gao2023addressing,bwgnn_tang2022rethinking,bahonar2019graph}. \textit{Thirdly}, the residual operation has also been shown to be effective in shallow GAD methods (e.g., Radar~\cite{li2017radar}), further supporting its utility in capturing anomaly-related features. Unlike shallow methods, RGE can adaptively learn transformations and capture higher-order residual information through the multi-hop residual design, which enhances its modeling capacity and enables it to better capture complex anomaly patterns. 

Once the residual representations are obtained at multiple iterations, RGE generates the final representations by concatenating them, which can be written by:

\begin{equation}
\label{eq:concat}
 \mathbf{H} = [\mathbf{R}^{[1]}||\cdots||\mathbf{R}^{[L]}], 
\end{equation}

\noindent where $\mathbf{H} \in \mathbb{R}^{n \times d_e}$ is the representation matrix and $||$ is the concatenation operator. By combining the residual information from different propagation steps, RGE can learn more comprehensive and anomaly-sensitive representations that capture patterns from diverse receptive fields. This is particularly beneficial for generalist GAD problems, as the anomaly patterns in datasets from different domains may span varying ranges of neighbors. 

\subsection{Cross-Attentive In-Context Anomaly Scoring}\label{subsec:scoring}

As the final stage of the pipeline of \ourmethod, the cross-attentive in-context anomaly scoring module aims to predict the abnormality (i.e., anomaly scores) of each unlabeled node based on the few-shot normal nodes. Unlike existing supervised or unsupervised GAD scenarios, generalist GAD scenarios require the model to make direct predictions during the testing phase based solely on the provided few-shot normal samples, without the need for dataset-specific fine-tuning or additional training. In this context, how to effectively leverage the few-shot normal nodes becomes crucial for making accurate predictions. 

To address this challenge, we propose a novel scoring module with a cross-attention mechanism for anomaly prediction based on the \textit{in-context learning paradigm}. Following this paradigm, we denote the few-shot normal nodes as \textbf{context nodes} $\mathcal{V}_k$, while the remaining unlabeled nodes are denoted as \textbf{query nodes} $\mathcal{V}_q$ that need to be predicted as anomalous or normal. In this setup, we split the representation matrix $\mathbf{H}\in \mathbb{R}^{n \times d_e}$ into two parts by indexing the corresponding row vectors: the representations of context nodes $\mathbf{H}_k \in \mathbb{R}^{n_k \times d_e}$ and the representations of query nodes $\mathbf{H}_q \in \mathbb{R}^{n_q \times d_e}$. Note that the number of context nodes $n_k$ can be much smaller than the number of query nodes $n_q$ in the few-shot setting. 

The core idea of our in-context learning module is to reconstruct each query node representation with the combination of context node representations. Then, the reconstruction error of each query node can serve as its anomaly score to measure its abnormality. To this end, we design a value-free cross-attention block for reconstruction, where each row vector of $\mathbf{H}_q$ (i.e., the representation of each query node) is represented through a linear combination of $\mathbf{H}_k$:

\begin{equation}
\label{eq:cross_attention}
\begin{aligned}
\mathbf{Q} &= \mathbf{H}_q \mathbf{W}_q, \quad \mathbf{K} = \mathbf{H}_k \mathbf{W}_k, \\
\tilde{\mathbf{H}}_q &= \operatorname{Softmax}\left(\frac{\mathbf{Q}\mathbf{K}^\top}{\sqrt{d_e}}\right) \mathbf{H}_k.
\end{aligned}
\end{equation}

\noindent where $\mathbf{Q} \in \mathbb{R}^{n_q \times d_e}$ and $\mathbf{K} \in \mathbb{R}^{n_k \times d_e}$ are the query and key matrices respectively, $\mathbf{W}_q$ and $\mathbf{W}_k$ are learnable parameters, and $\tilde{\mathbf{H}}_q\in \mathbb{R}^{n_q \times d_e}$ is the reconstructed query embedding matrix. Different from the cross-attention blocks in most models~\cite{vaswani2017attention,rombach2022high}, the value-free cross-attention block in \ourmethod directly multiplies the attention matrix with $\mathbf{H}_k$, without mapping $\mathbf{H}_k$ with an extra value matrix. In this way, we can ensure that $\tilde{\mathbf{H}}_q$ lies in the same representation space as ${\mathbf{H}}_q$, allowing the cross-attention block to perform reconstruction effectively. Given a query node $v_i$, we can predict its anomaly score $f(v_i)$ by calculating the L2 distance between its query representation vector ${\mathbf{H}}{q_i}$ and the corresponding reconstructed query representation vector ${\tilde{\mathbf{H}}q}_i$, which can be written by:

\begin{equation}
\label{eq:score}
f(v_i) = d({\mathbf{H}}{q_i}, \tilde{\mathbf{H}}{q_i}) = \sqrt{\sum_{j=1}^{d_e}\left({\mathbf{H}}{q_{ij}}-\tilde{\mathbf{H}}{q_{ij}}\right)^2}.
\end{equation}

\begin{figure}[!t]
  \centering
  \includegraphics[width=1\columnwidth]{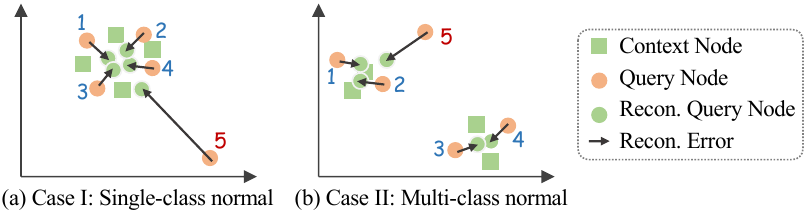}
  \caption{Toy examples of reconstruction-based scoring mechanism. In each case, nodes 1, 2, 3, and 4 are normal samples, while node 5 is an anomaly.}
  \label{fig:method_example}
\end{figure}

\noindent\textbf{Mechanism Analysis.} 
The design of the in-context anomaly scoring module is based on an assumption: The normal query nodes exhibit similar patterns to several context nodes, due to the inherent shared attributes within normal samples. As a result, their representations can be easily reconstructed by a linear combination of the context node representations. That is, given a normal node, its original and reconstructed representations can be close to each other in the representation space, leading to a smaller anomaly score $f(v_i)$. Conversely, since anomalies usually exhibit patterns that differ from those of normal nodes, it would not be easy to construct them using contextual normal nodes. Therefore, their reconstruction errors may be relevantly larger, resulting in higher anomaly scores.

To further illustrate the reconstruction-based scoring mechanism, we present two toy examples demonstrating how the reconstruction error functions as a measure of abnormality in different GAD scenarios. As shown in Fig.~\ref{fig:method_example}, we consider two typical scenarios: Case I - single-class normal, where all normal samples exhibit similar patterns or characteristics, and Case II - multi-class normal, where normal samples are grouped into multiple classes, each characterized by distinct patterns or traits. In Case I (Fig.~\ref{fig:method_example}(a)), we can witness that the normal query nodes stay in the same cluster with the context nodes, leading to their significantly smaller reconstruction errors compared to the abnormal node. Interestingly, if the cross-attention block assigns uniform weights to all context nodes, our scoring module can function as a one-class classification model~\cite{ruff2018deep}, where the average of the context node representations is the one-class centroid. This inherent similarity between our scoring module and traditional anomaly detection methods ensures compatibility and facilitates knowledge transfer across diverse anomaly detection scenarios. The toy example of Case II (Fig.~\ref{fig:method_example}(b)) further illustrates that our cross-attentive scoring module effectively handles the multi-class normal scenario. Specifically, given a normal query node, the cross-attention block can automatically identify and utilize the closest context samples belonging to the same cluster for reconstruction. 

\noindent\textbf{Model Training. } 
To train \ourmethod in an end-to-end manner, we utilize a marginal cosine similarity loss to minimize the reconstruction errors of normal nodes, while maximizing the reconstruction errors of abnormal nodes. In practice, for a training graph dataset with anomaly labels, we randomly select $n_k$ normal nodes as context nodes and sample an equal number of normal and abnormal nodes to serve as query nodes. Given a query node $v_i$ with representation ${\mathbf{h}}{q_i}$, reconstructed representation $\tilde{\mathbf{h}}{q_i}$, and anomaly label~${y}_i$, the sample-level loss function can be denoted as:

\begin{equation}
\label{eq:lossfunc}
\mathcal{L}= \begin{cases}1-\cos \left({\mathbf{h}}{q_i}, \tilde{\mathbf{h}}{q_i}\right), & \text { if } \mathbf{y}_i=0 \\ \max \left(0, \cos \left({\mathbf{h}}{q_i}, \tilde{\mathbf{h}}{q_i}\right)-\epsilon\right), & \text { if } \mathbf{y}_i=1\end{cases}
\end{equation}

\noindent where $\cos(\cdot,\cdot)$ and $\max(\cdot,\cdot)$ denote the cosine similarity and maximum operation and $\epsilon$ is a margin hyperparameter.

\section{ARC\textsubscript{zero}: Towards Zero-Shot Generalist GAD}
In the previous section, we propose \ourmethod, an effective few-shot generalist GAD approach. Nevertheless, in certain specific application domains where labeling costs are high or expert knowledge is required, even a few-shot set of normal samples may not be available. To extend generalist GAD to these domains, a zero-shot generalist GAD method capable of detecting anomalies without relying on labeled samples becomes highly desirable. As we already have a strong few-shot method \ourmethod, a natural question arises: \textit{can we develop a zero-shot generalist GAD method based on a well-trained \ourmethod model with minor specialized modifications?} 

To apply \ourmethod to the zero-shot setting, a promising approach is to select appropriate unlabeled nodes to serve as context nodes, making the entire process function effectively. Building on this idea, we propose \ourzero, a zero-shot generalist GAD method that leverages a pseudo-context mechanism for zero-shot anomaly scoring. The pipeline of \ourmethod is demonstrated by (a)\ding{220}(b)\ding{220}(d) in Fig.~\ref{fig:pipeline}. Using the same alignment module, encoder, and training process as \ourmethod, \ourzero introduces a new pseudo-context anomaly scoring module that self-adaptively selects pseudo-normal nodes to facilitate zero-shot anomaly detection. In the first step, we utilize a clustering algorithm to initialize the pseudo labels of few-shot normal nodes. Then, we iteratively execute the cross-attentive scoring module using the pseudo labels as input and then select new pseudo-context nodes from the remaining non-pseudo nodes according to the estimated scores. To conduct stable and reliable zero-shot anomaly detection, we aggregate the results from multiple iterations and compute the final anomaly scores by synthesizing the multi-round predictions. The following subsections provide a detailed explanation of each component. 

\subsection{Pseudo-Context Initialization}

The core idea of \ourzero is to leverage pseudo-normal nodes to approximate context nodes for effective zero-shot generalist GAD. In this context, the first step is to initialize the pseudo-normal nodes, serving as the cold-start context for the subsequent process. To select the pseudo-normal nodes as the context samples, we consider two key principles: \textit{representativeness} and \textit{diversity}. Specifically, the former requires the selected nodes to reflect the overall distribution of normal samples, while the latter indicates that the pseudo-normal nodes can capture a wide range of variations within the data. 

Following the above principles, we propose a clustering-based strategy for pseudo-context initialization. To be more concrete, a $k$-means clustering algorithm~\cite{hartigan1979k} is first applied to the aligned features of the target dataset, partitioning the nodes into $k$ clusters, each represented by a centroid. In practice, we set the number of clusters equal to the number of context nodes, i.e., $k=n_k$. \rev{For the choice of $n_k$, we can monitor self-consistency proxies, such as the convergence of iterative pseudo-context refinement and the tail-to-body separation of anomaly scores, to make better $n_k$ choices without performance validation.} After clustering, we initialize the pseudo-normal nodes as the samples closest to the centroids of each cluster.  The selected $n_k$ nodes can serve as the context nodes in the cross-attentive in-context anomaly scoring module at the first zero-shot scoring iteration.

Despite the simplicity of our clustering-based initialization strategy, it adheres to our principles of {representativeness} and {diversity}. On the one hand, the selected nodes are located near the cluster centers, where data distribution density is usually high, thus ensuring their \textit{representativeness} by capturing the dominant patterns within the dataset. On the other hand, the clustering-based initialization strategy ensures \textit{diversity} by selecting nodes from different clusters, with each cluster typically representing distinct patterns or normal categories, thereby capturing a broader spectrum of variations in the dataset. \rev{Moreover, even in extreme cases where frequent anomalies are mistakenly included in the initial pseudo-context, the subsequent iterative refinement and multi-round aggregation can progressively mitigate the impact of this imperfect start.} 

\subsection{Iterative Pseudo-Context Anomaly Scoring}

Taking the initialized pseudo-normal nodes as input, the cross-attentive scoring module in \ourmethod can now predict the abnormality of all nodes in a zero-shot manner. Nevertheless, the pseudo-normal nodes selected by the clustering algorithm may not adequately characterize the distribution of normal data; more critically, the selected nodes can even be noisy, \rev{i.e., anomaly nodes might be mistakenly selected as context nodes}, which further compromises the reliability of the predictions. In order to generate more reliable anomaly scores, we adopt a multi-iteration approach to refine the pseudo-normal nodes and generate more reliable zero-shot GAD predictions. In each iteration, we utilize the pseudo-normal nodes to compute anomaly scores for the current step; these scores are then employed to select more representative pseudo-normal nodes for the next iteration. 

Concretely, at the $t$-th iteration step, given a set of pseudo-normal nodes, we can separate the node set $\mathcal{V}$ into pseudo context node set $\mathcal{V}_k^{[t]}$ and query node set $\mathcal{V}_q^{[t]} = \mathcal{V} \setminus \mathcal{V}_k^{[t]}$. Following the procedure in Section~\ref{subsec:scoring}, the anomaly score $f(v_i)^{[t]}$ of each node $v_i \in \mathcal{V}_q^{[t]}$ can be estimated, which measures the abnormality of all the query nodes at the current step. After that, we can sample the ``most normal nodes'' from the query set as the new pseudo-normal nodes for the next iteration, enabling a refined representation of normal patterns for improved scoring in the following steps. In practice, we pick the nodes with top-$k$ smallest anomaly scores as the new pseudo-normal nodes, which can be written by:

\begin{equation}
\mathcal{V}_k^{[t+1]} = \operatorname{argmin}_{v_i \in \mathcal{V}_q^{[t]}}^{k} \{f(v_i)^{[t]}\},
\end{equation}

\noindent where $\mathcal{V}_k^{[t+1]}$ can play the role of pseudo-normal (context) nodes in the next step (i.e. the $(t+1)$-th iteration). By employing this iterative mechanism, the model refines the pseudo-normal nodes and anomaly predictions in a self-adaptive manner, improving the accuracy and robustness of the detection process.

\subsection{Multi-Round Anomaly Score Estimation}

After multiple iterations, the abnormality predictions based on different groups of refined pseudo-normal nodes can be obtained. A naive solution is to employ the scores at the final iteration as the final zero-shot GAD prediction. However, the GAD predictions from a single iteration are often sub-optimal and unreliable due to the selection bias of pseudo-normal nodes and the prediction bias inherent in zero-shot anomaly scoring. \rev{To mitigate the potential instability arising from imperfect pseudo-context initialization}, in \ourzero, we propose aggregating the anomaly scores across multiple iterations as the final output, collectively enhancing the reliability and robustness of anomaly detection. 
For the anomaly scores at each iteration, we first impute the scores of pseudo-normal nodes by assigning them the smallest score value from the current iteration. Formally, given $v_i \in \mathcal{V}_k^{[t]}$, we define:

\begin{equation}
f(v_i)^{[t]} = \operatorname{min}_{v_j \in \mathcal{V}_q^{[t]}} \{f(v_j)^{[t]}\}.
\end{equation}

\noindent The final anomaly score for each node $v_i \in \mathcal{V}$ is obtained by calculating the average score across all iterations:

\begin{equation}
f(v_i) = \frac{1}{T} \sum_{t=1}^{T} f(v_i)^{[t]}.
\end{equation}
\rev{The algorithm and complexity analysis of \ourzero see Appendices C and B, respectively.}

\section{Experiments}
\begin{table}[t]
\caption{The statistics of datasets.} 
\vspace{-2mm}
\centering
\label{tab:dset}
\resizebox{1\columnwidth}{!}{
\begin{tabular}{l|cc|cccc}
\toprule
Dataset & $\mathcal{T}_{train}$ & $\mathcal{T}_{test}$ & \#Nodes & \#Edges & \#Feat. & \#Ano. \\
\midrule
\rowcolor{dgreen}
\multicolumn{7}{c}{Citation network with injected anomalies} \\
Cora & - & $\checkmark$ & 2,708 & 5,429 & 1,433 & 150 \\
CiteSeer & - & $\checkmark$ & 3,327 & 4,732 & 3,703 & 150 \\
ACM & - & $\checkmark$ & 16,484 & 71,980 & 8,337 & 597 \\
PubMed & $\checkmark$ & - & 19,717 & 44,338 & 500 & 600 \\

\midrule
\rowcolor{dgreen}
\multicolumn{7}{c}{Social network with injected anomalies} \\
BlogCatalog & - & $\checkmark$ & 5,196 & 171,743 & 8,189 & 300 \\
Flickr & $\checkmark$ & - & 7,575 & 239,738 & 12,047 & 450 \\

\midrule
\rowcolor{dgreen}
\multicolumn{7}{c}{Social network with real anomalies} \\
Facebook & - & $\checkmark$ & 1,081 & 55,104 & 576 & 25 \\
Weibo & - & $\checkmark$ & 8,405 & 407,963 & 400 & 868 \\
Reddit & - & $\checkmark$ & 10,984 & 168,016 & 64 & 366 \\
Questions & $\checkmark$ & - & 48,921 & 153,540 & 301 & 1,460 \\

\midrule
\rowcolor{dgreen}
\multicolumn{7}{c}{Co-review network with real anomalies} \\
Amazon & - & $\checkmark$ & 10,244 & 175,608 & 25 & 693 \\
YelpChi & $\checkmark$ & - & 23,831 & 49,315 & 32 & 1,217 \\

\midrule
\rowcolor{dred}
\multicolumn{7}{c}{Out-of-domain datasets with injected anomalies} \\
Amazon-Photo & - & $\checkmark$ & 7,650 & 238,162 & 745 & 450 \\
Coauthor-CS & - & $\checkmark$ & 18,333 & 163,788 & 6,805 & 600 \\
\midrule
\rowcolor{dred}
\multicolumn{7}{c}{Out-of-domain datasets with real anomalies} \\
Tolokers & - & $\checkmark$ & 11,758 & 519,000 & 10 & 2,566 \\
T-Finance & - & $\checkmark$ & 39,357 & 21,222,543 & 10 & 1,803 \\
Elliptic & - & $\checkmark$ & 203,769 & 234,355 & 166 & 4,545 \\
\bottomrule
\end{tabular}

}
\end{table}

\begin{table*}[t]
\caption{Anomaly detection performance in terms of AUROC (in percent, mean$\pm$std). Highlighted are the results ranked \textcolor{firstcolor}{\textbf{\underline{first}}}, \textcolor{secondcolor}{\underline{second}}, and \textcolor{thirdcolor}{\underline{third}}. ``Rank'' indicates the average ranking over 13 datasets. Superscript ``*'' on a baseline method name indicates that our method is significantly different from this baseline on \texttt{ALL\_DATASETS\_TRIAL\_AVG} (paired t-test and Wilcoxon test, p$<$0.05).} 
\vspace{-2mm}
\centering
\label{tab:main_auroc}
\resizebox{1\textwidth}{!}{
\begin{tabular}{l|ccccccc}
\toprule
Method & Cora & CiteSeer & ACM & BlogCatalog & Facebook & Weibo & Reddit\\
\midrule
\rowcolor{Gray}
        \multicolumn{8}{c}{Supervised - Pre-Train Only} \\
GCN & $59.64{\scriptstyle\pm8.30}$ & $60.27{\scriptstyle\pm8.11}$ & $60.49{\scriptstyle\pm9.65}$ & $56.19{\scriptstyle\pm6.39}$ & $29.51{\scriptstyle\pm4.86}$ & $76.64{\scriptstyle\pm17.69}$ & $50.43{\scriptstyle\pm4.41}$ \\
GAT & $50.06{\scriptstyle\pm2.65}$ & $51.59{\scriptstyle\pm3.49}$ & $48.79{\scriptstyle\pm2.73}$ & $50.40{\scriptstyle\pm2.80}$ & $51.88{\scriptstyle\pm2.16}$ & $53.06{\scriptstyle\pm7.48}$ & $51.78{\scriptstyle\pm4.04}$ \\
BGNN & $42.45{\scriptstyle\pm11.57}$ & $42.32{\scriptstyle\pm11.82}$ & $44.00{\scriptstyle\pm13.69}$ & $47.67{\scriptstyle\pm8.52}$ & $54.74{\scriptstyle\pm25.29}$ & $32.75{\scriptstyle\pm35.35}$ & $50.27{\scriptstyle\pm3.84}$ \\
BWGNN & $54.06{\scriptstyle\pm3.27}$ & $52.61{\scriptstyle\pm2.88}$ & $67.59{\scriptstyle\pm0.70}$ & $56.34{\scriptstyle\pm1.21}$ & $45.84{\scriptstyle\pm4.97}$ & $53.38{\scriptstyle\pm1.61}$ & $48.97{\scriptstyle\pm5.74}$ \\
GHRN & $59.89{\scriptstyle\pm6.57}$ & $56.04{\scriptstyle\pm9.19}$ & $55.65{\scriptstyle\pm6.37}$ & $57.64{\scriptstyle\pm3.48}$ & $44.81{\scriptstyle\pm8.06}$ & $51.87{\scriptstyle\pm14.18}$ & $46.22{\scriptstyle\pm2.33}$ \\
\midrule
\rowcolor{Gray}
        \multicolumn{8}{c}{Unsupervised - Pre-Train Only} \\
DOMINANT & $66.53{\scriptstyle\pm1.15}$ & $69.47{\scriptstyle\pm2.02}$ & $70.08{\scriptstyle\pm2.34}$ & $74.25{\scriptstyle\pm0.65}$ & $51.01{\scriptstyle\pm0.78}$ & $\first{92.88{\scriptstyle\pm0.32}}$ & $50.05{\scriptstyle\pm4.92}$ \\
CoLA & $63.29{\scriptstyle\pm8.88}$ & $62.84{\scriptstyle\pm9.52}$ & $66.85{\scriptstyle\pm4.43}$ & $50.04{\scriptstyle\pm3.25}$ & $12.99{\scriptstyle\pm11.68}$ & $16.27{\scriptstyle\pm5.64}$ & $52.81{\scriptstyle\pm6.69}$ \\
HCM-A & $54.28{\scriptstyle\pm4.73}$ & $48.12{\scriptstyle\pm6.80}$ & $53.70{\scriptstyle\pm4.64}$ & $55.31{\scriptstyle\pm0.57}$ & $35.44{\scriptstyle\pm13.97}$ & $65.52{\scriptstyle\pm12.58}$ & $48.79{\scriptstyle\pm2.75}$ \\
TAM & $62.02{\scriptstyle\pm2.39}$ & $72.27{\scriptstyle\pm0.83}$ & $74.43{\scriptstyle\pm1.59}$ & $49.86{\scriptstyle\pm0.73}$ & $65.88{\scriptstyle\pm6.66}$ & $71.54{\scriptstyle\pm0.18}$ & $55.43{\scriptstyle\pm0.33}$ \\
\midrule
\rowcolor{Gray}
        \multicolumn{8}{c}{Unsupervised - Pre-Train \& Fine-Tune} \\
DOMINANT & $\third{72.23{\scriptstyle\pm0.34}}$ & $74.69{\scriptstyle\pm0.32}$ & $74.34{\scriptstyle\pm0.12}$ & $\third{74.61{\scriptstyle\pm0.04}}$ & $49.92{\scriptstyle\pm0.55}$ & $\second{92.21{\scriptstyle\pm0.10}}$ & $52.14{\scriptstyle\pm5.06}$ \\
CoLA & $67.62{\scriptstyle\pm4.26}$ & $70.75{\scriptstyle\pm3.42}$ & $69.11{\scriptstyle\pm0.67}$ & $62.49{\scriptstyle\pm3.38}$ & $64.70{\scriptstyle\pm18.86}$ & $31.55{\scriptstyle\pm6.02}$ & $\third{58.12{\scriptstyle\pm0.67}}$ \\
HCM-A & $56.45{\scriptstyle\pm4.93}$ & $55.54{\scriptstyle\pm4.07}$ & $57.69{\scriptstyle\pm3.59}$ & $55.10{\scriptstyle\pm0.29}$ & $36.57{\scriptstyle\pm10.72}$ & $71.89{\scriptstyle\pm2.79}$ & $49.15{\scriptstyle\pm2.72}$ \\
TAM & $62.56{\scriptstyle\pm2.10}$ & $\third{76.54{\scriptstyle\pm1.33}}$ & $\first{86.29{\scriptstyle\pm1.57}}$ & $57.69{\scriptstyle\pm0.88}$ & $\third{76.26{\scriptstyle\pm3.70}}$ & $71.73{\scriptstyle\pm0.16}$ & $56.62{\scriptstyle\pm0.49}$ \\

\rowcolor{Gray}
        \multicolumn{8}{c}{Generalist} \\
UNPrompt$^{*}$ & $67.02{\scriptstyle\pm3.06}$ & $72.84{\scriptstyle\pm1.95}$ & $74.93{\scriptstyle\pm0.84}$ & $68.80{\scriptstyle\pm0.18}$ & $\second{78.46{\scriptstyle\pm4.91}}$ & $48.52{\scriptstyle\pm0.86}$ & $56.75{\scriptstyle\pm0.84}$ \\
AnomalyGFM$^{*}$ & $50.67{\scriptstyle\pm2.90}$ & $50.63{\scriptstyle\pm3.07}$ & $51.13{\scriptstyle\pm2.53}$ & $45.89{\scriptstyle\pm0.39}$ & $\first{87.58{\scriptstyle\pm3.41}}$ & $66.31{\scriptstyle\pm8.12}$ & $55.93{\scriptstyle\pm3.62}$ \\
\ourmethod (ours) & $\second{87.45{\scriptstyle\pm0.74}}$ & $\second{90.95{\scriptstyle\pm0.59}}$ & $\second{79.88{\scriptstyle\pm0.28}}$ & $\first{74.76{\scriptstyle\pm0.06}}$ & $67.56{\scriptstyle\pm1.60}$ & $\third{88.85{\scriptstyle\pm0.14}}$ & $\first{60.04{\scriptstyle\pm0.69}}$ \\ 
\ourzero (ours) & $\first{87.54{\scriptstyle\pm0.57}}$ & $\first{91.20{\scriptstyle\pm0.27}}$ & $\third{78.78{\scriptstyle\pm0.17}}$ & $\second{74.75{\scriptstyle\pm0.03}}$ & $69.25{\scriptstyle\pm0.95}$ & $88.71{\scriptstyle\pm0.18}$ & $\second{59.57{\scriptstyle\pm0.42}}$ \\ 
\bottomrule
\toprule

Method & Amazon & Amazon-Photo & Coauthor-CS & Tolokers & T-Finance & Elliptic & Avg. Rank \\
\midrule
\rowcolor{Gray}
        \multicolumn{8}{c}{Supervised - Pre-Train Only} \\
GCN & $46.63{\scriptstyle\pm3.47}$ & $59.27{\scriptstyle\pm1.00}$ & $63.79{\scriptstyle\pm0.68}$ & $53.38{\scriptstyle\pm3.29}$ & $58.79{\scriptstyle\pm8.50}$ & $51.88{\scriptstyle\pm6.81}$ & $9.2$ \\
GAT & $50.52{\scriptstyle\pm17.22}$ & $52.80{\scriptstyle\pm10.08}$ & $50.16{\scriptstyle\pm3.38}$ & $46.23{\scriptstyle\pm11.95}$ & $45.05{\scriptstyle\pm14.44}$ & $\third{57.38{\scriptstyle\pm7.14}}$ & $12.1$\\
BGNN & $52.26{\scriptstyle\pm3.31}$ & $58.69{\scriptstyle\pm3.06}$ & $63.25{\scriptstyle\pm0.70}$ & $48.44{\scriptstyle\pm8.27}$ & $50.97{\scriptstyle\pm1.15}$ & $51.70{\scriptstyle\pm2.09}$ & $12.2$\\
BWGNN & $55.26{\scriptstyle\pm16.95}$ & $55.55{\scriptstyle\pm7.47}$ & $59.45{\scriptstyle\pm11.12}$ & $44.85{\scriptstyle\pm9.24}$ & $52.19{\scriptstyle\pm23.09}$ & $55.55{\scriptstyle\pm6.25}$ & $11.1$\\
GHRN & $49.48{\scriptstyle\pm17.13}$ & $56.18{\scriptstyle\pm1.83}$ & $68.64{\scriptstyle\pm0.83}$ & $\first{59.53{\scriptstyle\pm6.61}}$ & $54.43{\scriptstyle\pm10.62}$ & $54.39{\scriptstyle\pm5.70}$ & $9.8$\\
\midrule
\rowcolor{Gray}
        \multicolumn{8}{c}{Unsupervised - Pre-Train Only} \\
DOMINANT & $48.94{\scriptstyle\pm2.69}$ & $47.39{\scriptstyle\pm1.20}$ & $60.61{\scriptstyle\pm0.52}$ & $48.12{\scriptstyle\pm4.85}$ & OOM & OOM & $10.2$\\
CoLA & $47.40{\scriptstyle\pm7.97}$ & $56.08{\scriptstyle\pm0.15}$ & $70.26{\scriptstyle\pm0.39}$ & $52.36{\scriptstyle\pm2.02}$ & $51.80{\scriptstyle\pm0.15}$ & OOM & $10.8$\\
HCM-A & $43.99{\scriptstyle\pm0.72}$ & $52.38{\scriptstyle\pm0.64}$ & $41.69{\scriptstyle\pm1.75}$ & $51.35{\scriptstyle\pm0.85}$ & $54.84{\scriptstyle\pm1.56}$ & OOM & $13.0$ \\
TAM & $56.06{\scriptstyle\pm2.19}$ & $58.35{\scriptstyle\pm1.91}$ & $69.95{\scriptstyle\pm2.00}$ & $50.51{\scriptstyle\pm0.14}$ & OOM & OOM & $9.0$ \\
\midrule
\rowcolor{Gray}
        \multicolumn{8}{c}{Unsupervised - Pre-Train \& Fine-Tune} \\
DOMINANT & $59.06{\scriptstyle\pm2.80}$ & $60.60{\scriptstyle\pm0.11}$ & $73.63{\scriptstyle\pm0.16}$ & $47.44{\scriptstyle\pm0.64}$ & OOM & OOM & $7.3$ \\
CoLA & $52.51{\scriptstyle\pm6.66}$ & $64.21{\scriptstyle\pm0.43}$ & $\third{79.27{\scriptstyle\pm1.26}}$ & $\second{55.38{\scriptstyle\pm3.43}}$ & $23.81{\scriptstyle\pm1.36}$ & OOM & $7.2$\\
HCM-A & $42.20{\scriptstyle\pm0.55}$ & $52.86{\scriptstyle\pm1.38}$ & $45.25{\scriptstyle\pm3.86}$ & $49.59{\scriptstyle\pm3.67}$ & OOM & OOM & $12.8$\\
TAM & $57.13{\scriptstyle\pm1.59}$ & $57.28{\scriptstyle\pm2.73}$ & $69.96{\scriptstyle\pm2.01}$ & $50.81{\scriptstyle\pm0.03}$ & OOM & OOM & $7.2$\\

\rowcolor{Gray}
        \multicolumn{8}{c}{Generalist} \\
UNPrompt$^{*}$ & $61.62{\scriptstyle\pm5.07}$ & $\third{70.29{\scriptstyle\pm0.87}}$ & $73.87{\scriptstyle\pm1.47}$ & $45.17{\scriptstyle\pm1.54}$ & $20.47{\scriptstyle\pm2.26}$ & $52.83{\scriptstyle\pm1.46}$ & $\third{6.5}$\\
AnomalyGFM$^{*}$ & $\second{83.16{\scriptstyle\pm0.77}}$ & $52.17{\scriptstyle\pm0.98}$ & $44.54{\scriptstyle\pm2.77}$ & $\third{54.73{\scriptstyle\pm3.92}}$ & $\third{65.46{\scriptstyle\pm2.35}}$ & $40.93{\scriptstyle\pm5.69}$ & $9.8$\\
\ourmethod (ours) & $\third{80.67{\scriptstyle\pm1.81}}$ & $\first{75.86{\scriptstyle\pm0.71}}$ & $\second{82.29{\scriptstyle\pm0.59}}$ & $53.08{\scriptstyle\pm1.93}$ & $\second{73.88{\scriptstyle\pm3.43}}$ & $\first{79.33{\scriptstyle\pm0.49}}$ & $\first{2.3}$\\ 
\ourzero (ours) & $\first{83.76{\scriptstyle\pm1.01}}$ & $\second{75.30{\scriptstyle\pm0.92}}$ & $\first{82.67{\scriptstyle\pm0.10}}$ & $51.70{\scriptstyle\pm0.86}$ & $\first{75.47{\scriptstyle\pm0.31}}$ & $\second{78.84{\scriptstyle\pm0.12}}$ & $\second{2.4}$\\ 
\bottomrule
\end{tabular}
}
\vspace{-2mm}
\end{table*}

\subsection{Experimental Setup}

\subsubsection{Datasets}

For comprehensive evaluations, we consider graph datasets spanning multiple domains, such as social media, academic, e-commerce, crowd-sourcing service, and finance, with either injected anomalies or real anomalies~\cite{tang2024gadbench,cola_liu2021anomaly,tam_qiao2024truncated,shchur2018pitfalls}. We partition all datasets into a training set, $\mathcal{T}_{train}$, and a testing set, $\mathcal{T}_{test}$, ensuring there is no overlap between the two sets. Specifically, we first collect four basic groups of datasets: 1) citation network with injected anomalies, including Cora, CiteSeer, ACM, and PubMed~\cite{sen2008collective,tang2008arnetminer}; 2) social network with injected anomalies, including BlogCatalog and Flickr~\cite{tang2009relational}; 3) social network with real anomalies, including Facebook, Weibo, Reddit, and Questions~\cite{xu2022contrastive, kumar2019predicting, platonov2023critical}; 4) co-review network with real anomalies, including Amazon and YelpChi~\cite{rayana2015collective,mcauley2013amateurs}. Within each group, we consider the largest dataset to be one of the training datasets $\mathcal{T}_{train}$, and the rest are regarded as the in-domain testing datasets. To verify the out-of-domain generalization ability of the proposed methods, we consider 5 out-of-domain datasets: Coauthor-CS, a co-author network~\cite{shchur2018pitfalls}; Amazon-Photo, a co-purchase network~\cite{shchur2018pitfalls}; Tolokers, a crowd-sourcing service network~\cite{platonov2023critical}; T-Finance, a finance network~\cite{bwgnn_tang2022rethinking}; and Elliptic, a payment flow network~\cite{weber2019anti}. These out-of-domain datasets exhibit significant diversity compared to the training datasets, as they come from diverse, new, and unseen domains. The detailed statistics of the datasets are shown in Table~\ref{tab:dset}\rev{, with additional explanations provided in Appendix~A}.

\subsubsection{Baselines}

We compare the proposed methods with those previous state-of-the-art methods, which can be generally divided into two categories: supervised methods and unsupervised methods. The supervised baselines include two conventional GNNs, i.e., GCN~\cite{gcn_kipf2017semi} and GAT~\cite{gat_velivckovic2018graph}, and three state-of-the-art supervised GAD methods, i.e., BGNN~\cite{bgnn_ivanov2021boost}, BWGNN~\cite{bwgnn_tang2022rethinking}, and GHRN~\cite{ghrn_gao2023addressing}. The unsupervised methods comprise four representative approaches, each featuring a unique design, including the generative method DOMINANT~\cite{dominant_ding2019deep}, the contrastive method CoLA~\cite{cola_liu2021anomaly}, the hop predictive method HCM-A~\cite{huang2022hop}, and the affinity-based method TAM~\cite{tam_qiao2024truncated}. \rev{In addition, we also compare with two very recent generalist GAD methods, AnomalyGFM~\cite{qiao2025anomalygfm} and UNPrompt~\cite{niu2024zero}, to evaluate the performance against the latest state-of-the-art.}

\subsubsection{Evaluation Metrics}

Following previous studies~\cite{tang2024gadbench,tam_qiao2024truncated}, we employ two popular and complementary evaluation metrics for evaluation, including area under the receiver operating characteristic Curve (AUROC) and area under the precision-recall curve (AUPRC). A higher AUROC/AUPRC value indicates better performance. To ensure robustness, we report the average AUROC and AUPRC along with their standard deviations over 5 trials.

\subsubsection{Implementation Details}

To implement the generalist GAD settings, we train the baselines as well as the proposed methods on all the datasets in $\mathcal{T}_{train}$ and evaluate them on each dataset in $\mathcal{T}_{test}$. During the inference phase, \ourmethod performs predictions on each testing dataset using an in-context learning approach, with a default shot number of $n_k=10$. For \ourzero, we also set the number of pseudo-normal nodes as 10. For the supervised baselines, we train the models on $\mathcal{T}_{train}$ and directly evaluate on each $\mathcal{D}^{(i)}_{test} \in \mathcal{T}_{test}$, since no labeled anomaly is available for fine-tuning (denoted as ``pre-train only''). For the unsupervised baselines, we evaluate two settings: ``pre-train only'' and ``pre-train \& fine-tune''. In the latter setting, we perform additional dataset-specific fine-tuning for a few epochs. We standardized the testing samples for each dataset across different methods to ensure a fair comparison. To standardize the feature space in the baseline methods, we employ either a learnable projection or a random projection as an adapter between the raw features and the model's input layer. We use random search to optimize the hyperparameters for both the baselines and the proposed methods. We avoid dataset-specific hyperparameter tuning to align with our goal of developing generalist GAD models. Instead, we apply a single set of hyperparameters for all testing datasets\rev{, i.e., using one unified hyperparameter configuration across all datasets (training and testing) without any dataset-specific tuning}. \rev{Specifically, for \ourmethod and \ourzero, we fix the key hyperparameters (e.g., representation dimension $h=64$, layer number $L=2$, shot number $n_k=10$) across all datasets. For the baselines, we conduct 50 trials of random search within their specific parameter spaces to ensure optimal performance.}

\subsection{Performance Comparison}

The performance comparison on 13 testing datasets is summarized in Table~\ref{tab:main_auroc} in terms of AUROC. The corresponding AUPRC results are reported in Appendix~D. Based on the results, we have made the following observations regarding the proposed methods \ourmethod and \ourzero:

\begin{itemize}[leftmargin=*, itemsep=0pt, parsep=0pt, topsep=2pt]
    \item \textbf{Superior performance.} The proposed \ourmethod and \ourzero consistently outperform baseline methods across nearly all datasets, as shown by their significantly higher AUROC and AUPRC scores. This highlights their ability to generalize to diverse datasets under the generalist GAD scenarios, achieving state-of-the-art performance for both supervised and unsupervised baselines. \rev{Moreover, compared to the latest generalist baselines (UNPrompt and AnomalyGFM), \ourzero achieve consistently stronger overall performance, and the difference is statistically significant based on paired t-test and Wilcoxon signed-rank test (p$<$0.05).}
    \item \textbf{Less application costs.} Our methods achieve strong performance without requiring any dataset-specific fine-tuning, making them highly efficient and cost-effective for real-world applications. In contrast, the baseline methods under the ``pre-train only'' setting exhibit generally mediocre performance. Even when fine-tuning is applied -- at the cost of additional computational resources and manual effort -- these baselines often fail to match the performance of our methods. 
    \item \textbf{Strong out-of-domain generalization ability.} The proposed methods, \ourmethod and \ourzero, demonstrate excellent performance on five out-of-domain datasets, even when the training datasets come from entirely different domains. This highlights their ability to adapt to new and unseen domains without requiring domain-specific fine-tuning. Such robust cross-domain generalization is particularly valuable in real-world scenarios, where training data often differ significantly from the target application domains. 
    \item \textbf{Zero-shot detection capability of \ourzero.} Surprisingly, \ourzero achieves competitive performance, even without using any labeled data during inference. \rev{In some scenarios, \ourzero even outperforms \ourmethod. This is because its iterative pseudo-context refinement progressively selects highly representative normal nodes to improve context quality over potentially noisy fixed few-shot examples, thereby enabling reliable zero-shot anomaly detection on unseen datasets.}
    \item \textbf{Scalability to large datasets.} Both \ourmethod and \ourzero exhibit strong scalability, performing effectively on large datasets such as T-Finance and Elliptic. This scalability is further demonstrated by their consistent performance even under resource constraints, such as when some baseline methods run out of memory (OOM).
\end{itemize}

\subsection{Ablation Studies}

\begin{table*}[t]
\caption{Performance of \ourmethod and its variants in terms of AUROC.} 
\vspace{-2mm}
\centering
\label{tab:ablation_full_auroc}
\resizebox{1\textwidth}{!}{
\begin{tabular}{l|cccccccc}
\toprule
Variant & Cora & CiteSeer & ACM & BlogCatalog & Facebook & Weibo & Reddit & Amazon \\
\midrule
\ourmethod & $\first{87.45{\scriptstyle\pm0.74}}$ & $\first{90.95{\scriptstyle\pm0.59}}$ & $\first{79.88{\scriptstyle\pm0.28}}$ & $\first{74.76{\scriptstyle\pm0.06}}$ & $\first{67.56{\scriptstyle\pm1.60}}$ & $88.85{\scriptstyle\pm0.14}$ & $\first{60.04{\scriptstyle\pm0.69}}$ & $\first{80.67{\scriptstyle\pm1.81}}$ \\
\midrule
w/o A & $80.65{\scriptstyle\pm0.71}$ & $83.35{\scriptstyle\pm0.64}$ & $79.29{\scriptstyle\pm0.16}$ & $73.86{\scriptstyle\pm0.18}$ & $62.80{\scriptstyle\pm2.06}$ & $89.69{\scriptstyle\pm0.17}$ & $54.60{\scriptstyle\pm1.92}$ & $64.76{\scriptstyle\pm2.13}$ \\
w/o R & $37.44{\scriptstyle\pm1.40}$ & $31.52{\scriptstyle\pm0.71}$ & $61.83{\scriptstyle\pm1.16}$ & $49.30{\scriptstyle\pm2.06}$ & $20.38{\scriptstyle\pm9.63}$ & $\first{97.72{\scriptstyle\pm0.59}}$ & $52.94{\scriptstyle\pm0.96}$ & $50.15{\scriptstyle\pm0.24}$ \\
w/o C & $47.39{\scriptstyle\pm0.42}$ & $53.98{\scriptstyle\pm0.72}$ & $54.24{\scriptstyle\pm1.32}$ & $60.46{\scriptstyle\pm1.23}$ & $48.86{\scriptstyle\pm0.97}$ & $42.84{\scriptstyle\pm3.01}$ & $51.03{\scriptstyle\pm0.86}$ & $69.02{\scriptstyle\pm0.97}$ \\
\bottomrule
\end{tabular}
}
\vspace{-2mm}
\end{table*}



\begin{table*}[t]
\caption{Performance of \ourzero and its variants in terms of AUROC.} 
\vspace{-2mm}
\centering
\label{tab:ablation_zero_auroc}
\resizebox{1\textwidth}{!}{
\begin{tabular}{l|cccccccc}
\toprule
Variant & Cora & CiteSeer & ACM & BlogCatalog & Facebook & Weibo & Reddit & Amazon \\
\midrule
\ourzero & $87.54{\scriptstyle\pm0.57}$ & $91.20{\scriptstyle\pm0.27}$ & $78.78{\scriptstyle\pm0.17}$ & $\first{74.75{\scriptstyle\pm0.03}}$ & $\first{69.25{\scriptstyle\pm0.95}}$ & $\first{88.71{\scriptstyle\pm0.18}}$ & $\first{59.57{\scriptstyle\pm0.42}}$ & $\first{83.76{\scriptstyle\pm1.01}}$ \\
\midrule
w/o Iter. Score & $87.76{\scriptstyle\pm0.60}$ & $90.13{\scriptstyle\pm0.39}$ & $79.23{\scriptstyle\pm0.13}$ & $74.71{\scriptstyle\pm0.15}$ & $67.78{\scriptstyle\pm1.00}$ & $88.17{\scriptstyle\pm0.13}$ & $59.13{\scriptstyle\pm0.53}$ & $80.85{\scriptstyle\pm1.64}$ \\
w/o Multi-Round & $87.37{\scriptstyle\pm0.55}$ & $\first{91.48{\scriptstyle\pm0.26}}$ & $\first{79.70{\scriptstyle\pm0.09}}$ & $74.47{\scriptstyle\pm0.04}$ & $65.52{\scriptstyle\pm0.93}$ & $88.25{\scriptstyle\pm0.13}$ & $58.16{\scriptstyle\pm0.74}$ & $80.92{\scriptstyle\pm1.09}$ \\
\midrule
w Random Init & $86.47{\scriptstyle\pm0.17}$ & $90.10{\scriptstyle\pm0.21}$ & $79.69{\scriptstyle\pm0.04}$ & $74.15{\scriptstyle\pm0.13}$ & $65.40{\scriptstyle\pm1.12}$ & $79.36{\scriptstyle\pm4.32}$ & $58.92{\scriptstyle\pm0.33}$ & $82.91{\scriptstyle\pm0.27}$ \\
w Degree Init & $87.30{\scriptstyle\pm0.52}$ & $91.31{\scriptstyle\pm0.23}$ & $79.56{\scriptstyle\pm0.09}$ & $73.92{\scriptstyle\pm0.17}$ & $68.46{\scriptstyle\pm1.62}$ & $84.28{\scriptstyle\pm2.18}$ & $59.21{\scriptstyle\pm0.32}$ & $78.82{\scriptstyle\pm2.53}$ \\
w Rep. Clu. Init & $\first{87.87{\scriptstyle\pm0.53}}$ & $91.45{\scriptstyle\pm0.23}$ & $74.51{\scriptstyle\pm0.95}$ & $69.59{\scriptstyle\pm1.18}$ & $68.39{\scriptstyle\pm1.09}$ & $60.42{\scriptstyle\pm13.45}$ & $52.57{\scriptstyle\pm1.31}$ & $72.26{\scriptstyle\pm1.70}$ \\
\bottomrule
\end{tabular}
}
\vspace{-2mm}
\end{table*}

To verify the effectiveness of key components and designs in \ourmethod and \ourzero, we conduct ablation studies by removing or modifying specific modules and evaluating their impact on performance. For brevity, we present only the results in terms of AUROC in ablation studies.

\subsubsection{Ablation Study for \ourmethod}

To investigate the contributions of the three core components in \ourmethod, we replace each component with an alternative design and construct three variants: 1) \textbf{w/o A}: using random projection to replace smoothness-based feature alignment module; 2) \textbf{w/o R}: using GCN to replace ego-neighbor residual graph encoder; and 
3) \textbf{w/o C}: using binary classification-based predictor trained by binary cross-entropy loss to replace cross-attentive in-context anomaly scoring module. The results of the performance comparison are shown in Table~\ref{tab:ablation_full_auroc}. From the table, we can see that all three components significantly contribute to the performance, as removing or replacing any of them leads to a notable drop in GAD performance. Among these components, the in-context anomaly scoring module proves to be crucial, as removing it (\textbf{w/o C}) results in performance that is nearly equivalent to random guessing on the majority of datasets. The residual graph encoder also plays a significant role, as we observe that removing this component leads to the performance flip issue~\cite{zhao2023using} on certain datasets, indicating unreliable anomaly detection judgments. Notably, the Weibo dataset stands out as an exception, where the GCN encoder outperforms the residual graph encoder. This could be attributed to the unique anomaly patterns present in the Weibo dataset, which differ from those in the other datasets.

\subsubsection{Ablation Study for \ourzero}
For \ourmethod, we create two variants to discuss the impact of the key designs: 1) \textbf{w/o Iter. Score}: using single step scoring based on initialized pseudo-normal nodes to replace iterative scoring and 2) \textbf{w/o Multi-Round}: using the predictions at the final iteration as the output anomaly scores. Meanwhile, to investigate the effectiveness of different initialization strategies, here we try three alternative ways, including random selection, mean-degree nodes as initialization, and clustering based on representations. The comparison results are demonstrated in Table~\ref{tab:ablation_zero_auroc}, leading to the following observations. 
1) Both iterative scoring and multi-round refinement prove to be critical components of \ourzero. Removing iterative scoring results in consistently lower performance across datasets, highlighting the importance of iteratively refining the anomaly scores. Similarly, removing the multi-round process also degrades performance, particularly on datasets like Facebook and Reddit, where the iterative process aids in capturing complex patterns. 
2) Among the initialization strategies, random selection shows the weakest performance, indicating that initialization with meaningful prior knowledge is essential. Using degree-based initialization slightly improves performance over random selection, as it exploits structural information about the graph. However, the representation-based clustering as initialization sometimes yields stronger results across most datasets. 
3) Despite some variance in performance depending on the dataset, the full \ourzero method outperforms all variants in most cases, showcasing the robustness and effectiveness of its design choices. These results further validate the necessity of incorporating both iterative scoring and multi-round refinement, as well as using an feature clustering-based initialization strategy.

\subsection{Parameter Studies}

In this subsection, we analyze the impact of the number of context nodes (i.e., $n_k$) on the performance of \ourmethod and \ourzero, respectively. For brevity, we present the results on four representative in-domain datasets. 

\subsubsection{Effect of Few-Shot Sample Number in \ourmethod}
Firstly, we vary the number of few-shot normal nodes ($n_k$ in \ourmethod) within the range of 2 to 100 to evaluate the few-shot efficiency of \ourmethod. The results in terms of AUROC and AUPRC are illustrated in Fig. \ref{fig:shot_num_fs} (more results in Appendix~D). From the figure, we observe that \ourmethod consistently improves its performance as the number of few-shot normal nodes increases. Even with as few as $n_k = 6$ normal nodes, \ourmethod achieves competitive performance on most datasets. This demonstrates the effectiveness of our approach in leveraging limited labeled information. 
Notably, the performance gain tends to plateau beyond $n_k = 50$ for many datasets, indicating that \ourmethod efficiently utilizes the provided few-shot nodes without requiring a large number of labeled samples. This property makes \ourmethod particularly suitable for scenarios where available labeled data is costly. 

\begin{figure}[!t]
 \centering
 \subfigure[Cora]{
   \includegraphics[width=0.45\columnwidth]{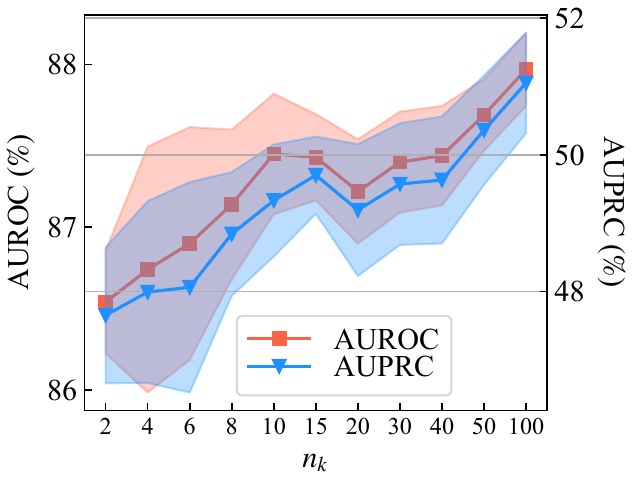}
   \label{subfig:shot_num_cora}
 } 
 \hfill
 \subfigure[Facebook]{
   \includegraphics[width=0.45\columnwidth]{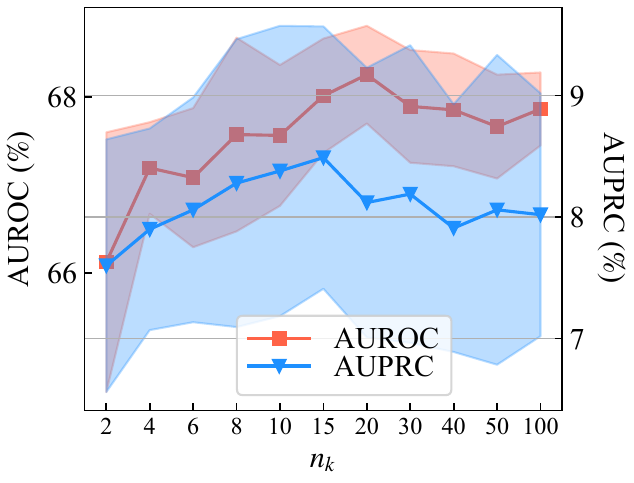}
   \label{subfig:shot_num_fb}
 } 
 \caption{Performance with varying $n_k$ in few-shot method \ourmethod. \rev{The shaded areas represent the standard deviation over 5 repeated runs.}}
 \vspace{-2mm}
 \label{fig:shot_num_fs}
\end{figure}

\begin{figure}[!t]
 \centering
 \subfigure[Cora]{
   \includegraphics[width=0.45\columnwidth]{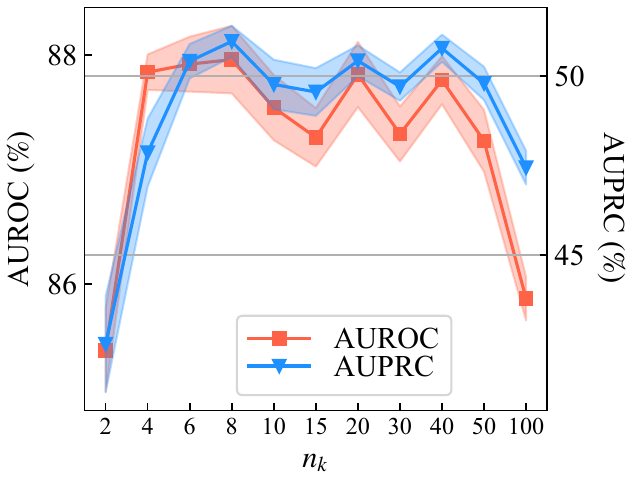}
   \label{subfig:shot_num_zs_cora}
 } 
 \hfill
 \subfigure[Facebook]{
   \includegraphics[width=0.45\columnwidth]{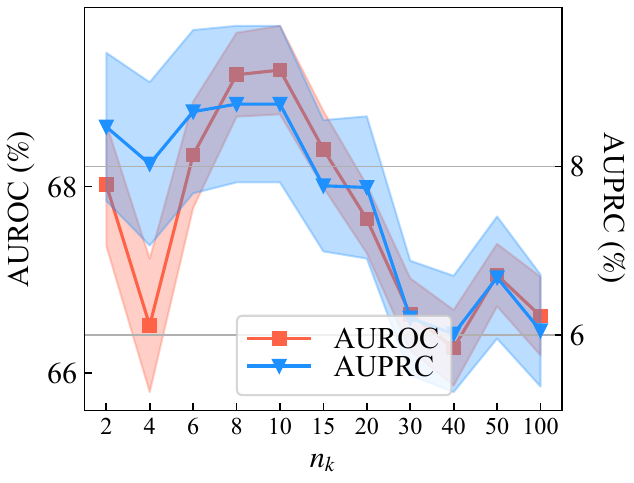}
   \label{subfig:shot_num_zs_fb}
 } 
 \vspace{-2mm}
 \caption{Performance with varying $n_k$ in zero-shot method \ourzero. \rev{The shaded areas represent the standard deviation over 5 repeated runs.}}
 \label{fig:shot_num_zs}
\end{figure} 

\subsubsection{Effect of Zero-Shot Pseudo-Normal Sample Number in \ourzero}
Unlike in the few-shot counterpart, in our zero-shot method \ourzero, $n_k$ represents the number of pseudo-normal samples, making it a flexible hyperparameter that can influence the model's performance. The experimental results in Fig.~\ref{fig:shot_num_zs} (more results in Appendix~D) reveal how varying $n_k$ from 2 to 100 impacts the zero-shot GAD performance. 
From the plots, it is evident that the performance of \ourzero is highly dependent on the choice of $n_k$. For some datasets (e.g. Cora), there is an optimal range of $n_k$ (often between 6 and 50) where both AUROC and AUPRC reach their peak. Beyond this range, performance either plateaus or deteriorates, likely due to the inclusion of less informative and noisy (abnormal) pseudo-normal samples that may introduce noise. 
Other datasets, such as Facebook and Amazon, exhibit a more erratic trend, with performance fluctuating as $n_k$ increases. This suggests that the quality and representativeness of pseudo-normal samples in these datasets are crucial, and a careful selection of $n_k$ is required to balance model stability and performance. 

\subsection{Efficiency Analysis}

\begin{figure}[!t]
  \centering
  \includegraphics[width=0.95\columnwidth]{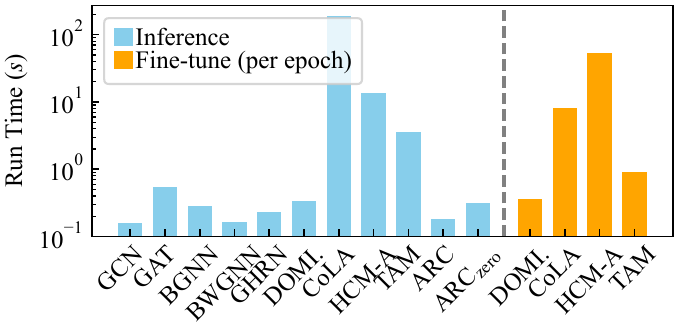}
  \vspace{-2mm}
  \caption{Efficiency comparison of inference and fine-tuning time.}
  \vspace{-2mm}
  \label{fig:efficiency}
\end{figure}

In this experiment, we compare the running efficiency of the proposed methods and baselines. Specifically, we compare the inference and fine-tuning times when applying the models to an unseen dataset, which is demonstrated in Fig. \ref{fig:efficiency}. 
The results demonstrate that \ourmethod achieves a fast inference speed comparable to the most efficient baselines, highlighting its practicality for real-world applications. Additionally, the zero-shot version, \ourzero, also exhibits relatively high efficiency, enabling zero-shot anomaly detection without the need for dataset-specific tuning. 
In contrast, dataset-specific fine-tuning imposes significant time costs. Even a single fine-tuning epoch for these baselines takes substantially longer than the inference phase of most methods, further underscoring the efficiency advantage of our generalist methods.

\subsection{Visualization}

\subsubsection{Visualization of Attention Weights}
\begin{figure}[t!]
\vspace{-1.9mm}
\centering
 \subfigure[Cora]{
   \includegraphics[width=0.28\columnwidth,angle=-90]{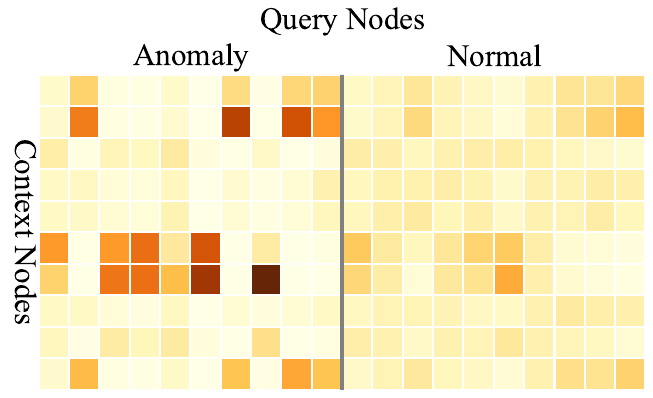}
   \label{subfig:attn_a}
}
 \hfill
 \subfigure[Amazon]{
   \includegraphics[width=0.28\columnwidth,angle=-90]{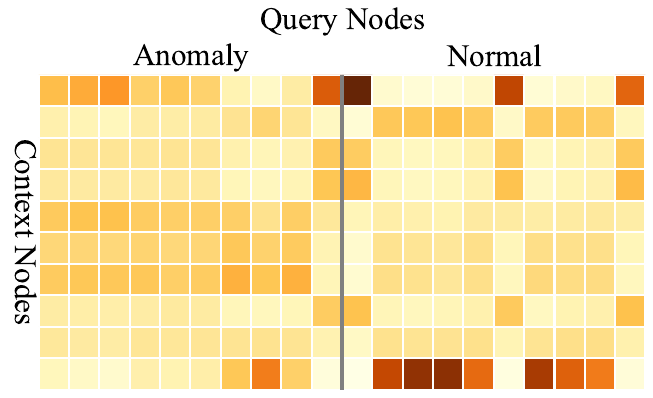}
   \label{subfig:attn_b}
}
\caption{Visualization of the attention weights of cross-attentive scoring module. Darker colors represent larger values.}
\label{fig:vis_attn}
\end{figure}

To investigate the weight allocation mechanism of the cross-attentive in-context anomaly scoring module in \ourmethod, we visualize the attention weights between context nodes and query nodes captured by the cross-attention block. The visualization results are illustrated in Fig. \ref{fig:vis_attn}. 
In Fig.~\ref{subfig:attn_a}, we observe that \ourmethod assigns relatively uniform attention weights to the normal nodes, leading to embeddings that closely mirror the average embedding of the surrounding context nodes. This indicates a more centralized representation of normal nodes. In contrast, anomalies are reconstructed using a selective combination of one or two context nodes, which results in embeddings that are more distanced from the central cluster. This behavior reflects the situation described in the ``single-class normal'' case in Fig.~\ref{fig:method_example}(a). 
On the other hand, Fig. \ref{subfig:attn_b} shows that each normal query node is allocated to multiple context nodes according to two distinct patterns, which is indicative of the ``multi-class normal'' scenario presented in Fig. \ref{fig:method_example}(b). This highlights the versatility of \ourmethod's cross-attention module, which facilitates adaptation to different anomaly and normal node distribution patterns. 

\begin{figure}[t!]
\vspace{-1.3mm}
\centering
 \subfigure[\ourmethod (ours)]{
   \includegraphics[width=0.14\textwidth]{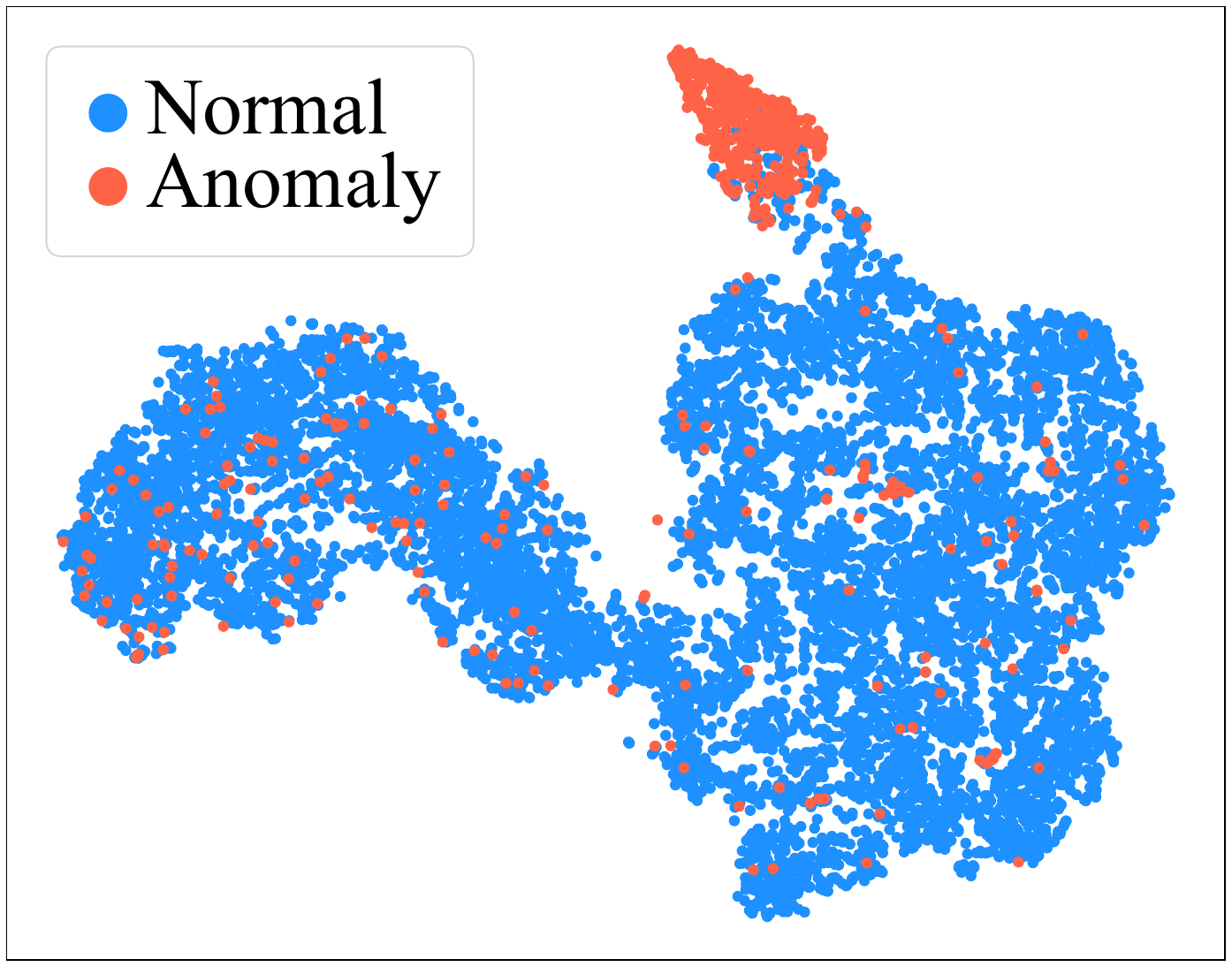}
}
 \hfill
 \subfigure[TAM]{
   \includegraphics[width=0.14\textwidth]{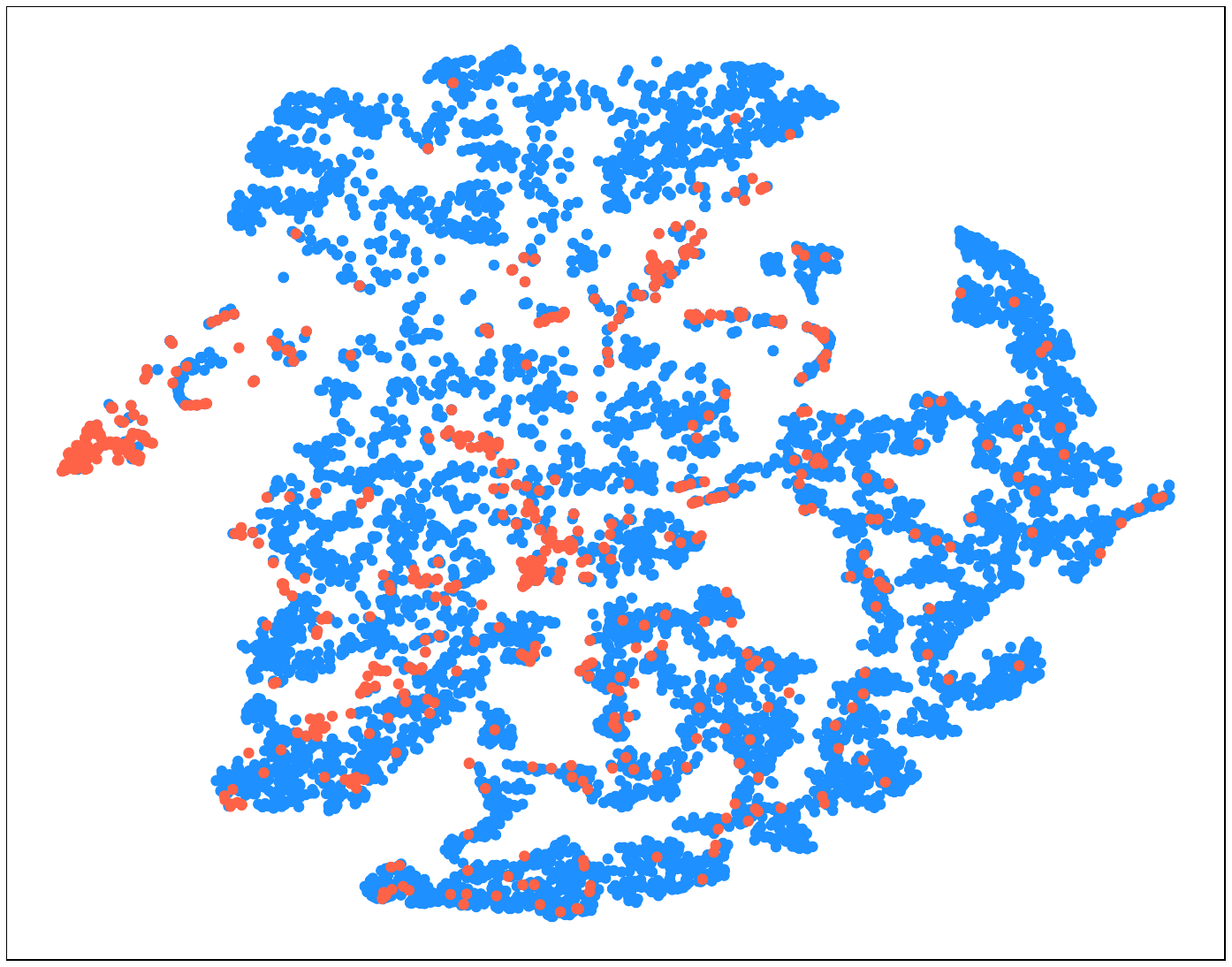}
}
 \hfill
 \subfigure[UNPrompt]{
   \includegraphics[width=0.14\textwidth]{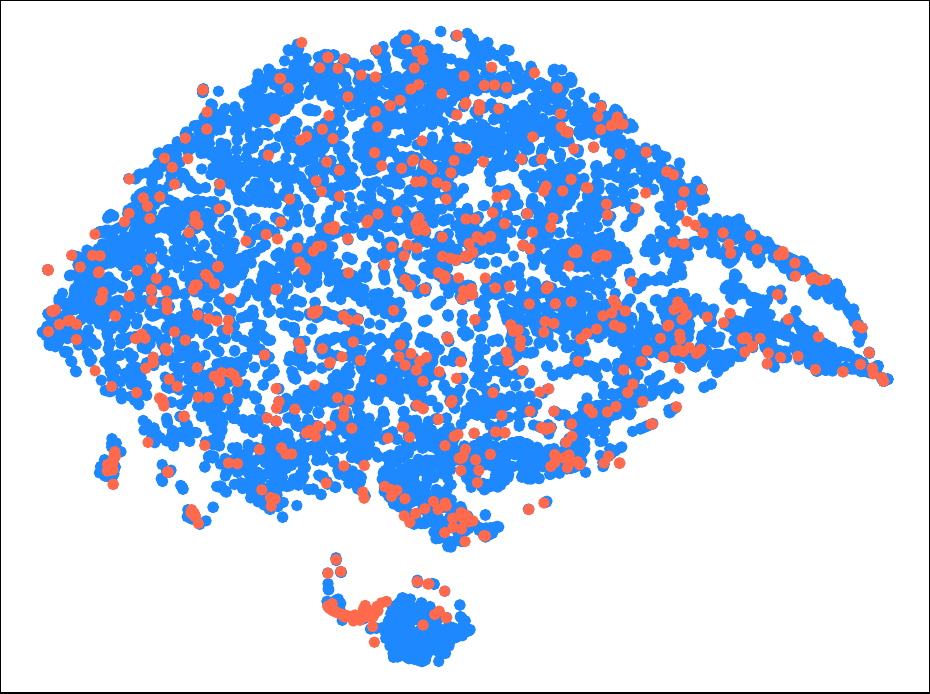}
}
\caption{t-SNE visualization of representations by \ourmethod and baselines.}
\label{fig:vis_tsne}
\end{figure}

\subsubsection{Visualization of Representations}
To investigate the representation learning ability of our ego-neighbor residual graph encoder, we visualize the representation distribution of Amazon, one of the testing datasets, learned by \ourmethod and three baselines (TAM and UNPrompt) using t-SNE~\cite{van2008visualizing_tsne}. The visualizations in Fig.~\ref{fig:vis_tsne} demonstrate the effectiveness of our encoder in generating discriminative representations for normal and anomalous nodes compared to the baseline. Specifically, we can observe that the anomalous samples are mainly clustered into one or more small groups, which enhances the separability between normal and anomalous nodes. This phenomenon suggests that the encoder is able to effectively capture the underlying patterns of anomalies, making them more distinguishable from the normal nodes. Notably, these visualizations are based on the representations of the test data, which were never seen during the training phase. This showcases the generalization capability of the model, as it successfully adapts to unseen data and effectively captures underlying patterns in the node distributions.

\section{Conclusion and Outlook}
While graph anomaly detection (GAD) has gained increasing research attention in recent years, the challenge of developing a universal and general GAD model that can be directly applied to any new dataset on-the-fly remains unresolved. This paper addresses this challenge by introducing the novel research problem of generalist GAD, which aims to develop models capable of detecting anomalies across diverse datasets without requiring task-specific fine-tuning. Firstly, we develop a few-shot generalist GAD method, termed \ourmethod, which leverages in-context learning to effectively detect anomalies on an unseen dataset using only a few labeled normal samples. Building on \ourmethod, we further propose a zero-shot variant, \ourzero, which eliminates the need for labeled samples by generating pseudo-normal nodes for anomaly detection, enabling fully label-free generalization to unseen datasets. Extensive experiments conducted on diverse benchmark datasets demonstrate the effectiveness and efficiency of both \ourmethod and \ourzero, highlighting their ability to achieve competitive GAD performance across various domains and scenarios. 

\noindent\textbf{Limitations.} 
Despite its promising performance, our study has some limitations that open avenues for future research. \textbf{Limitation 1:} In certain scenarios, \ourmethod may be outperformed by \ourzero. A possible explanation is that \ourzero applies a filtering mechanism to context samples, which enhances the quality of the pseudo-normal nodes. 
\textbf{Limitation 2:} In \ourzero, determining the optimal number of pseudo-normal nodes and ensuring that the selected nodes align well with the underlying data distribution remain open challenges. \rev{If the initial selection is not ideal, there is a risk of confirmation bias being amplified during iterations,} which could affect the model's stability and generalization on unseen datasets. \rev{\textbf{Limitation 3:} From a theoretical perspective, providing formal guarantees for the proposed generalist GAD framework remains an open problem.}

\noindent\textbf{Future Directions.} 
In future research, several promising directions can be explored. 1) To handle \textbf{Limitation 1}, potential solutions include developing semi-supervised in-context learning frameworks that leverage both labeled and unlabeled samples or employing active learning strategies~\cite{chen2022supervised} that allow the model to select the most informative nodes to label. 2) To deal with \textbf{Limitation 2}, adaptive mechanisms can be developed to dynamically optimize the definition and selection of pseudo-normal nodes. \rev{For instance, one could add lightweight consistency or diversity constraints during pseudo-context selection (e.g., filtering nodes by cross-run stability) to further mitigate the impact of imperfect initialization and ensure robustness across varying data distributions.} \rev{3) While this work introduces the concept of generalist GAD, our focus is limited to node-level anomaly detection. Extending this concept to edge-level and graph-level GAD, or designing a universal model capable of accommodating anomalies at multiple scales, represents an exciting and promising direction for future exploration.} \rev{4) Although we focus on universal generalization without adaptation, investigating target-specific fine-tuning remains an interesting avenue to further probe the robustness of the learned parameter space.}

\section*{Acknowledgment}
The work of S. Pan was partially supported by ARC Grant No. DP240101547. The work of Y. Liu was partially supported by the ARC Grant No. DE260101172.

\bibliographystyle{IEEEtran}
\bibliography{ref}

@article{sen2008collective,
  title={Collective classification in network data},
  author={Sen, Prithviraj and Namata, Galileo and Bilgic, Mustafa and Getoor, Lise and Galligher, Brian and Eliassi-Rad, Tina},
  journal={AI magazine},
  volume={29},
  number={3},
  pages={93--93},
  year={2008}
}

@inproceedings{tam_qiao2024truncated,
  title={Truncated affinity maximization: One-class homophily modeling for graph anomaly detection},
  author={Qiao, Hezhe and Pang, Guansong},
  booktitle={Advances in Neural Information Processing Systems},
  volume={36},
  year={2023}
}

@inproceedings{ruff2018deep,
  title={Deep one-class classification},
  author={Ruff, Lukas and Vandermeulen, Robert and Goernitz, Nico and Deecke, Lucas and Siddiqui, Shoaib Ahmed and Binder, Alexander and M{\"u}ller, Emmanuel and Kloft, Marius},
  booktitle={International conference on machine learning},
  pages={4393--4402},
  year={2018},
  organization={PMLR}
}

@inproceedings{roth2022towards,
  title={Towards total recall in industrial anomaly detection},
  author={Roth, Karsten and Pemula, Latha and Zepeda, Joaquin and Sch{\"o}lkopf, Bernhard and Brox, Thomas and Gehler, Peter},
  booktitle={Proceedings of the IEEE/CVF Conference on Computer Vision and Pattern Recognition},
  pages={14318--14328},
  year={2022}
}

@inproceedings{defard2021padim,
  title={Padim: a patch distribution modeling framework for anomaly detection and localization},
  author={Defard, Thomas and Setkov, Aleksandr and Loesch, Angelique and Audigier, Romaric},
  booktitle={International Conference on Pattern Recognition},
  pages={475--489},
  year={2021},
  organization={Springer}
}

@inproceedings{zhou2017anomaly,
  title={Anomaly detection with robust deep autoencoders},
  author={Zhou, Chong and Paffenroth, Randy C},
  booktitle={Proceedings of the 23rd ACM SIGKDD international conference on knowledge discovery and data mining},
  pages={665--674},
  year={2017}
}

@inproceedings{sehwag2021ssd,
  title={{SSD}: A unified framework for self-supervised outlier detection},
  author={Sehwag, Vikash and Chiang, Mung and Mittal, Prateek},
  booktitle={International Conference on Learning Representations},
  year={2021}
}

@inproceedings{jeong2023winclip,
  title={Winclip: Zero-/few-shot anomaly classification and segmentation},
  author={Jeong, Jongheon and Zou, Yang and Kim, Taewan and Zhang, Dongqing and Ravichandran, Avinash and Dabeer, Onkar},
  booktitle={Proceedings of the IEEE/CVF Conference on Computer Vision and Pattern Recognition},
  pages={19606--19616},
  year={2023}
}

@inproceedings{inctrl_zhu2024toward,
  title={Toward generalist anomaly detection via in-context residual learning with few-shot sample prompts},
  author={Zhu, Jiawen and Pang, Guansong},
  booktitle={Proceedings of the IEEE/CVF Conference on Computer Vision and Pattern Recognition},
  year={2024}
}

@inproceedings{dominant_ding2019deep,
  title={Deep anomaly detection on attributed networks},
  author={Ding, Kaize and Li, Jundong and Bhanushali, Rohit and Liu, Huan},
  booktitle={Proceedings of the 2019 SIAM International Conference on Data Mining},
  pages={594--602},
  year={2019},
  organization={SIAM}
}

@article{ma2021comprehensive,
  title={A comprehensive survey on graph anomaly detection with deep learning},
  author={Ma, Xiaoxiao and Wu, Jia and Xue, Shan and Yang, Jian and Zhou, Chuan and Sheng, Quan Z and Xiong, Hui and Akoglu, Leman},
  journal={IEEE Transactions on Knowledge and Data Engineering},
  volume={35},
  number={12},
  pages={12012--12038},
  year={2021},
  publisher={IEEE}
}

@inproceedings{bwgnn_tang2022rethinking,
  title={Rethinking graph neural networks for anomaly detection},
  author={Tang, Jianheng and Li, Jiajin and Gao, Ziqi and Li, Jia},
  booktitle={International Conference on Machine Learning},
  pages={21076--21089},
  year={2022},
  organization={PMLR}
}

@article{zhao2023using,
  title={On using classification datasets to evaluate graph outlier detection: Peculiar observations and new insights},
  author={Zhao, Lingxiao and Akoglu, Leman},
  journal={Big Data},
  volume={11},
  number={3},
  pages={151--180},
  year={2023},
  publisher={Mary Ann Liebert, Inc., publishers 140 Huguenot Street, 3rd Floor New~…}
}

@article{tang2024gadbench,
  title={Gadbench: Revisiting and benchmarking supervised graph anomaly detection},
  author={Tang, Jianheng and Hua, Fengrui and Gao, Ziqi and Zhao, Peilin and Li, Jia},
  journal={Advances in Neural Information Processing Systems},
  volume={36},
  year={2024}
}

@inproceedings{caregnn_dou2020enhancing,
  title={Enhancing graph neural network-based fraud detectors against camouflaged fraudsters},
  author={Dou, Yingtong and Liu, Zhiwei and Sun, Li and Deng, Yutong and Peng, Hao and Yu, Philip S},
  booktitle={Proceedings of the 29th ACM international conference on information \& knowledge management},
  pages={315--324},
  year={2020}
}

@inproceedings{ghrn_gao2023addressing,
  title={Addressing heterophily in graph anomaly detection: A perspective of graph spectrum},
  author={Gao, Yuan and Wang, Xiang and He, Xiangnan and Liu, Zhenguang and Feng, Huamin and Zhang, Yongdong},
  booktitle={Proceedings of the ACM Web Conference 2023},
  pages={1528--1538},
  year={2023}
}

@article{he2021bernnet,
  title={Bernnet: Learning arbitrary graph spectral filters via bernstein approximation},
  author={He, Mingguo and Wei, Zhewei and Xu, Hongteng and others},
  journal={Advances in Neural Information Processing Systems},
  volume={34},
  pages={14239--14251},
  year={2021}
}

@inproceedings{fan2020anomalydae,
  title={Anomalydae: Dual autoencoder for anomaly detection on attributed networks},
  author={Fan, Haoyi and Zhang, Fengbin and Li, Zuoyong},
  booktitle={ICASSP 2020-2020 IEEE International Conference on Acoustics, Speech and Signal Processing (ICASSP)},
  pages={5685--5689},
  year={2020},
  organization={IEEE}
}

@article{cola_liu2021anomaly,
  title={Anomaly detection on attributed networks via contrastive self-supervised learning},
  author={Liu, Yixin and Li, Zhao and Pan, Shirui and Gong, Chen and Zhou, Chuan and Karypis, George},
  journal={IEEE transactions on neural networks and learning systems},
  volume={33},
  number={6},
  pages={2378--2392},
  year={2021},
  publisher={IEEE}
}

@inproceedings{huang2022hop,
  title={Hop-count based self-supervised anomaly detection on attributed networks},
  author={Huang, Tianjin and Pei, Yulong and Menkovski, Vlado and Pechenizkiy, Mykola},
  booktitle={Joint European conference on machine learning and knowledge discovery in databases},
  pages={225--241},
  year={2022},
  organization={Springer}
}

@article{ding2021cross,
  title={Cross-domain graph anomaly detection},
  author={Ding, Kaize and Shu, Kai and Shan, Xuan and Li, Jundong and Liu, Huan},
  journal={IEEE Transactions on Neural Networks and Learning Systems},
  volume={33},
  number={6},
  pages={2406--2415},
  year={2021},
  publisher={IEEE}
}

@inproceedings{wang2023cross,
  title={Cross-domain graph anomaly detection via anomaly-aware contrastive alignment},
  author={Wang, Qizhou and Pang, Guansong and Salehi, Mahsa and Buntine, Wray and Leckie, Christopher},
  booktitle={Proceedings of the AAAI Conference on Artificial Intelligence},
  volume={37},
  pages={4676--4684},
  year={2023}
}

@article{unilp_dong2024universal,
  title={Universal link predictor by In-context Learning},
  author={Dong, Kaiwen and Mao, Haitao and Guo, Zhichun and Chawla, Nitesh V},
  journal={arXiv preprint arXiv:2402.07738},
  year={2024}
}

@inproceedings{gcn_kipf2017semi,
  title={Semi-supervised classification with graph convolutional networks},
  author={Kipf, Thomas N and Welling, Max},
  booktitle={International Conference on Learning Representations},
  year={2017}
}

@inproceedings{gat_velivckovic2018graph,
  title={Graph attention networks},
  author={Veli{\v{c}}kovi{\'c}, Petar and Cucurull, Guillem and Casanova, Arantxa and Romero, Adriana and Lio, Pietro and Bengio, Yoshua},
  booktitle={International Conference on Learning Representations},
  year={2018}
}

@inproceedings{bgnn_ivanov2021boost,
  title={Boost then convolve: Gradient boosting meets graph neural networks},
  author={Ivanov, Sergei and Prokhorenkova, Liudmila},
  booktitle={International Conference on Learning Representations},
  year={2021}
}

@article{pca_abdi2010principal,
  title={Principal component analysis},
  author={Abdi, Herv{\'e} and Williams, Lynne J},
  journal={Wiley interdisciplinary reviews: computational statistics},
  volume={2},
  number={4},
  pages={433--459},
  year={2010},
  publisher={Wiley Online Library}
}

@inproceedings{gcope_zhao2024all,
  title={All in one and one for all: A simple yet effective method towards cross-domain graph pretraining},
  author={Zhao, Haihong and Chen, Aochuan and Sun, Xiangguo and Cheng, Hong and Li, Jia},
  booktitle={Proceedings of the 30th ACM SIGKDD Conference on Knowledge Discovery and Data Mining},
  pages={4443--4454},
  year={2024}
}

@inproceedings{liu2024arc,
  title={ARC: A Generalist Graph Anomaly Detector with In-Context Learning},
  author={Liu, Yixin and Li, Shiyuan and Zheng, Yu and Chen, Qingfeng and Zhang, Chengqi and Pan, Shirui},
  booktitle={Advances in Neural Information Processing Systems},
  year={2024}
}

@inproceedings{ofa_liu2023one,
  title={One for all: Towards training one graph model for all classification tasks},
  author={Liu, Hao and Feng, Jiarui and Kong, Lecheng and Liang, Ningyue and Tao, Dacheng and Chen, Yixin and Zhang, Muhan},
  booktitle={International Conference on Learning Representations},
  year={2024}
}

@inproceedings{dong2021adagnn,
  title={Adagnn: Graph neural networks with adaptive frequency response filter},
  author={Dong, Yushun and Ding, Kaize and Jalaian, Brian and Ji, Shuiwang and Li, Jundong},
  booktitle={Proceedings of the 30th ACM international conference on information \& knowledge management},
  pages={392--401},
  year={2021}
}

@inproceedings{li2017radar,
  title={Radar: Residual analysis for anomaly detection in attributed networks.},
  author={Li, Jundong and Dani, Harsh and Hu, Xia and Liu, Huan},
  booktitle={IJCAI},
  volume={17},
  pages={2152--2158},
  year={2017}
}

@inproceedings{peng2018anomalous,
  title={ANOMALOUS: A Joint Modeling Approach for Anomaly Detection on Attributed Networks.},
  author={Peng, Zhen and Luo, Minnan and Li, Jundong and Liu, Huan and Zheng, Qinghua and others},
  booktitle={IJCAI},
  volume={18},
  pages={3513--3519},
  year={2018}
}

@inproceedings{sgc_wu2019simplifying,
  title={Simplifying graph convolutional networks},
  author={Wu, Felix and Souza, Amauri and Zhang, Tianyi and Fifty, Christopher and Yu, Tao and Weinberger, Kilian},
  booktitle={International conference on machine learning},
  pages={6861--6871},
  year={2019},
  organization={PMLR}
}

@article{vaswani2017attention,
  title={Attention is all you need},
  author={Vaswani, Ashish and Shazeer, Noam and Parmar, Niki and Uszkoreit, Jakob and Jones, Llion and Gomez, Aidan N and Kaiser, {\L}ukasz and Polosukhin, Illia},
  journal={Advances in neural information processing systems},
  volume={30},
  year={2017}
}

@inproceedings{rombach2022high,
  title={High-resolution image synthesis with latent diffusion models},
  author={Rombach, Robin and Blattmann, Andreas and Lorenz, Dominik and Esser, Patrick and Ommer, Bj{\"o}rn},
  booktitle={Proceedings of the IEEE/CVF conference on computer vision and pattern recognition},
  pages={10684--10695},
  year={2022}
}

@article{huang2024prodigy,
  title={Prodigy: Enabling in-context learning over graphs},
  author={Huang, Qian and Ren, Hongyu and Chen, Peng and Kr{\v{z}}manc, Gregor and Zeng, Daniel and Liang, Percy S and Leskovec, Jure},
  journal={Advances in Neural Information Processing Systems},
  volume={36},
  year={2024}
}

@inproceedings{tang2008arnetminer,
  title={Arnetminer: extraction and mining of academic social networks},
  author={Tang, Jie and Zhang, Jing and Yao, Limin and Li, Juanzi and Zhang, Li and Su, Zhong},
  booktitle={Proceedings of the 14th ACM SIGKDD international conference on Knowledge discovery and data mining},
  pages={990--998},
  year={2008}
}

@inproceedings{tang2009relational,
  title={Relational learning via latent social dimensions},
  author={Tang, Lei and Liu, Huan},
  booktitle={Proceedings of the 15th ACM SIGKDD international conference on Knowledge discovery and data mining},
  pages={817--826},
  year={2009}
}

@inproceedings{mcauley2013amateurs,
  title={From amateurs to connoisseurs: modeling the evolution of user expertise through online reviews},
  author={McAuley, Julian John and Leskovec, Jure},
  booktitle={Proceedings of the 22nd international conference on World Wide Web},
  pages={897--908},
  year={2013}
}

@inproceedings{rayana2015collective,
  title={Collective opinion spam detection: Bridging review networks and metadata},
  author={Rayana, Shebuti and Akoglu, Leman},
  booktitle={Proceedings of the 21th acm sigkdd international conference on knowledge discovery and data mining},
  pages={985--994},
  year={2015}
}

@inproceedings{xu2022contrastive,
  title={Contrastive attributed network anomaly detection with data augmentation},
  author={Xu, Zhiming and Huang, Xiao and Zhao, Yue and Dong, Yushun and Li, Jundong},
  booktitle={Pacific-Asia Conference on Knowledge Discovery and Data Mining},
  pages={444--457},
  year={2022},
  organization={Springer}
}

@inproceedings{kumar2019predicting,
  title={Predicting dynamic embedding trajectory in temporal interaction networks},
  author={Kumar, Srijan and Zhang, Xikun and Leskovec, Jure},
  booktitle={Proceedings of the 25th ACM SIGKDD international conference on knowledge discovery \& data mining},
  pages={1269--1278},
  year={2019}
}

@inproceedings{platonov2023critical,
  title={A critical look at the evaluation of GNNs under heterophily: Are we really making progress?},
  author={Platonov, Oleg and Kuznedelev, Denis and Diskin, Michael and Babenko, Artem and Prokhorenkova, Liudmila},
  booktitle = {Proceedings of the Twelfth International Conference on Learning Representations},
  year={2023}
}

@article{alayrac2022flamingo,
  title={Flamingo: a visual language model for few-shot learning},
  author={Alayrac, Jean-Baptiste and Donahue, Jeff and Luc, Pauline and Miech, Antoine and Barr, Iain and Hasson, Yana and Lenc, Karel and Mensch, Arthur and Millican, Katherine and Reynolds, Malcolm and others},
  journal={Advances in neural information processing systems},
  volume={35},
  pages={23716--23736},
  year={2022}
}

@article{bar2022visual,
  title={Visual prompting via image inpainting},
  author={Bar, Amir and Gandelsman, Yossi and Darrell, Trevor and Globerson, Amir and Efros, Alexei},
  journal={Advances in Neural Information Processing Systems},
  volume={35},
  pages={25005--25017},
  year={2022}
}

@inproceedings{pan2023prem,
  title={PREM: A Simple Yet Effective Approach for Node-Level Graph Anomaly Detection},
  author={Pan, Junjun and Liu, Yixin and Zheng, Yizhen and Pan, Shirui},
  booktitle={2023 IEEE International Conference on Data Mining (ICDM)},
  pages={1253--1258},
  year={2023},
  organization={IEEE}
}

@article{zheng2022graph,
  title={Graph neural networks for graphs with heterophily: A survey},
  author={Zheng, Xin and Wang, Yi and Liu, Yixin and Li, Ming and Zhang, Miao and Jin, Di and Yu, Philip S and Pan, Shirui},
  journal={arXiv preprint arXiv:2202.07082},
  year={2022}
}

@article{ahmed2021graph,
  title={Graph regularized autoencoder and its application in unsupervised anomaly detection},
  author={Ahmed, Imtiaz and Galoppo, Travis and Hu, Xia and Ding, Yu},
  journal={IEEE transactions on pattern analysis and machine intelligence},
  volume={44},
  number={8},
  pages={4110--4124},
  year={2021},
  publisher={IEEE}
}

@article{su2024large,
  title={Large language models for forecasting and anomaly detection: A systematic literature review},
  author={Su, Jing and Jiang, Chufeng and Jin, Xin and Qiao, Yuxin and Xiao, Tingsong and Ma, Hongda and Wei, Rong and Jing, Zhi and Xu, Jiajun and Lin, Junhong},
  journal={arXiv preprint arXiv:2402.10350},
  year={2024}
}

@inproceedings{zhou2023anomalyclip,
  title={Anomalyclip: Object-agnostic prompt learning for zero-shot anomaly detection},
  author={Zhou, Qihang and Pang, Guansong and Tian, Yu and He, Shibo and Chen, Jiming},
  booktitle={International Conference on Learning Representations},
  year={2024}
}

@article{qiao2024deep,
  title={Deep graph anomaly detection: A survey and new perspectives},
  author={Qiao, Hezhe and Tong, Hanghang and An, Bo and King, Irwin and Aggarwal, Charu and Pang, Guansong},
  journal={arXiv preprint arXiv:2409.09957},
  year={2024}
}

@inproceedings{dong2022survey,
  title={A survey on in-context learning},
  author={Dong, Qingxiu and Li, Lei and Dai, Damai and Zheng, Ce and Ma, Jingyuan and Li, Rui and Xia, Heming and Xu, Jingjing and Wu, Zhiyong and Chang, Baobao and others},
  booktitle={Proceedings of the 2024 Conference on Empirical Methods in Natural Language Processing},
  pages={1107--1128},
  year={2024}
}

@article{wang2022wrongdoing,
  title={Wrongdoing monitor: A graph-based behavioral anomaly detection in cyber security},
  author={Wang, Cheng and Zhu, Hangyu},
  journal={IEEE Transactions on Information Forensics and Security},
  volume={17},
  pages={2703--2718},
  year={2022},
  publisher={IEEE}
}

@article{wu2021graph,
  title={Graph neural networks for anomaly detection in industrial Internet of Things},
  author={Wu, Yulei and Dai, Hong-Ning and Tang, Haina},
  journal={IEEE Internet of Things Journal},
  volume={9},
  number={12},
  pages={9214--9231},
  year={2021},
  publisher={IEEE}
}

@article{hartigan1979k,
  title={A k-means clustering algorithm},
  author={Hartigan, John A and Wong, Manchek A and others},
  journal={Applied statistics},
  volume={28},
  number={1},
  pages={100--108},
  year={1979},
  publisher={USA}
}

@inproceedings{shchur2018pitfalls,
  title={Pitfalls of graph neural network evaluation},
  author={Shchur, Oleksandr and Mumme, Maximilian and Bojchevski, Aleksandar and G{\"u}nnemann, Stephan},
  booktitle={NeurIPS Workshop},
  year={2018}
}

@inproceedings{weber2019anti,
  title={Anti-money laundering in bitcoin: Experimenting with graph convolutional networks for financial forensics},
  author={Weber, Mark and Domeniconi, Giacomo and Chen, Jie and Weidele, Daniel Karl I and Bellei, Claudio and Robinson, Tom and Leiserson, Charles E},
  booktitle={KDD Workshop},
  year={2019}
}

@article{van2008visualizing_tsne,
  title={Visualizing data using t-SNE.},
  author={Van der Maaten, Laurens and Hinton, Geoffrey},
  journal={Journal of machine learning research},
  volume={9},
  number={11},
  year={2008}
}

@article{chen2022supervised,
  title={Supervised anomaly detection via conditional generative adversarial network and ensemble active learning},
  author={Chen, Zhi and Duan, Jiang and Kang, Li and Qiu, Guoping},
  journal={IEEE Transactions on Pattern Analysis and Machine Intelligence},
  volume={45},
  number={6},
  pages={7781--7798},
  year={2022},
  publisher={IEEE}
}

@article{xiang2024exploiting,
  title={Exploiting Structural Consistency of Chest Anatomy for Unsupervised Anomaly Detection in Radiography Images},
  author={Xiang, Tiange and Zhang, Yixiao and Lu, Yongyi and Yuille, Alan and Zhang, Chaoyi and Cai, Weidong and Zhou, Zongwei},
  journal={IEEE Transactions on Pattern Analysis and Machine Intelligence},
  year={2024},
  publisher={IEEE}
}

@article{bahonar2019graph,
  title={Graph embedding using frequency filtering},
  author={Bahonar, Hoda and Mirzaei, Abdolreza and Sadri, Saeed and Wilson, Richard C},
  journal={IEEE transactions on pattern analysis and machine intelligence},
  volume={43},
  number={2},
  pages={473--484},
  year={2019},
  publisher={IEEE}
}

@inproceedings{qiao2025anomalygfm,
  title={Anomalygfm: Graph foundation model for zero/few-shot anomaly detection},
  author={Qiao, Hezhe and Niu, Chaoxi and Chen, Ling and Pang, Guansong},
  booktitle={Proceedings of the 31st ACM SIGKDD Conference on Knowledge Discovery and Data Mining V. 2},
  pages={2326--2337},
  year={2025}
}

@inproceedings{niu2024zero,
  title={Zero-shot generalist graph anomaly detection with unified neighborhood prompts},
  author={Niu, Chaoxi and Qiao, Hezhe and Chen, Changlu and Chen, Ling and Pang, Guansong},
  booktitle = {Proceedings of the International Joint Conference on Artificial Intelligence},
  year={2025},
}

@inproceedings{zhao2025freegad,
  title={Freegad: A training-free yet effective approach for graph anomaly detection},
  author={Zhao, Yunfeng and Liu, Yixin and Li, Shiyuan and Chen, Qingfeng and Zheng, Yu and Pan, Shirui},
  booktitle={Proceedings of the 34th ACM International Conference on Information and Knowledge Management},
  pages={4379--4389},
  year={2025}
}

@inproceedings{pan2026correcting,
  title={Correcting False Alarms from Unseen: Adapting Graph Anomaly Detectors at Test Time},
  author={Pan, Junjun and Liu, Yixin and Zhou, Chuan and Xiong, Fei and Liew, Alan Wee-Chung and Pan, Shirui},
  booktitle={Proceedings of the AAAI Conference on Artificial Intelligence},
  year={2026}
}

@article{miao2025blindguard,
  title={Blindguard: Safeguarding llm-based multi-agent systems under unknown attacks},
  author={Miao, Rui and Liu, Yixin and Wang, Yili and Shen, Xu and Tan, Yue and Dai, Yiwei and Pan, Shirui and Wang, Xin},
  journal={arXiv preprint arXiv:2508.08127},
  year={2025}
}

@inproceedings{pan2025survey,
  title={A Survey of Generalization of Graph Anomaly Detection: From Transfer Learning to Foundation Models},
  author={Pan, Junjun and Zheng, Yu and Tan, Yue and Liu, Yixin},
  booktitle={The 16th IEEE International Conference on Knowledge Graphs},
  year={2025}
}

\clearpage

\appendices
\section{Additional Explanations for Datasets with Atypical Results}
\label{app:anomaly_datasets}

We provide additional explanations for the datasets used in our experiments, including their anomaly definitions and characteristics.
\begin{itemize}
    \item \textbf{Cora, CiteSeer, PubMed~\cite{sen2008collective}, and ACM~\cite{tang2008arnetminer}} are citation networks, where nodes are publications and edges denote citation links. Node attributes are bag-of-words features.
    \item \textbf{BlogCatalog and Flickr~\cite{dominant_ding2019deep,tang2009relational}} are social networks with injected anomalies, where nodes are users and edges represent following relationships. Node attributes are derived from user-generated textual content (e.g., posts, tags).
    \item \textbf{Amazon and YelpChi~\cite{rayana2015collective,mcauley2013amateurs}} are review-related graphs with real anomalies. In this work, we use Amazon-UPU and YelpChi-RUR, following the standard constructions based on user and review relations~\cite{tam_qiao2024truncated}.
    \item \textbf{Facebook~\cite{xu2022contrastive}} is a social network with real anomalies, where nodes are users and edges represent friendship relationships.
    \item \textbf{Reddit~\cite{kumar2019predicting}} is a forum interaction network, where banned users are treated as anomalies. Node attributes are derived from textual content.
    \item \textbf{Weibo~\cite{kumar2019predicting}} is a microblog user-hashtag graph, where users with frequent suspicious posting behaviors in a short time window are labeled as anomalies. Node features include location and bag-of-words features.
    \item \textbf{Questions~\cite{platonov2023critical}} is a Q\&A interaction network from Yandex Q, where nodes are users and edges indicate question-answer interactions. Node features are based on FastText embeddings of user descriptions.
    \item \textbf{Tolokers~\cite{platonov2023critical}} is a crowd-sourcing worker graph from the Toloka platform, where nodes are workers and edges connect workers who have worked on the same task. The task is to identify banned workers, and node features are based on worker profiles and task performance statistics.
    \item \textbf{T-Finance~\cite{bwgnn_tang2022rethinking}} is a transaction network, where nodes are anonymized accounts with 10-dimensional features (e.g., registration days, login activities, and interaction frequency) and edges indicate transaction records between accounts. Anomalies are accounts annotated by experts (e.g., fraud, money laundering, and online gambling).
    \item \textbf{Elliptic~\cite{weber2019anti}} is a Bitcoin transaction graph with directed payment flows, where nodes are transactions and edges denote payments. The nodes are associated with licit entities (e.g., exchanges and wallet providers) or illicit categories (e.g., scams, malware, ransomware, and Ponzi schemes), and anomalies correspond to illicit transactions.
\end{itemize}

\noindent\textbf{Anomaly Injection.} For datasets with injected anomalies, we follow the common protocol in~\cite{dominant_ding2019deep,cola_liu2021anomaly} to inject both structural and attribute anomalies.

\section{Complexity Analysis}
\label{app:complexity}
We summarize the time complexity of the overall inference pipeline (including \ourzero in Alg.~\ref{alg:arczero_infer_algo}) by decomposing it into feature alignment, embedding generation, and anomaly scoring.

In the testing phase, the time complexity consists of two main components: feature alignment and model inference. For feature alignment, the overall complexity is $\mathcal{O}(ndd_u + d_um + d_ulog(d_u))$, where $m = |\mathcal{E}|$ is the number of edges. Here, the first term is used for feature projection, while the second and third terms are used for smoothness computation and feature reordering, respectively. The model inference is divided into two main parts: embedding generation and anomaly scoring. The complexity of node embedding generation is $\mathcal{O}(L(md_u + nd_uh + nh^2))$, where the first term is used for feature propagation and the rest of the terms are used for residual encoding by MLP. The anomaly scoring, on the other hand, mainly involves cross-attention computation with time complexity of $\mathcal{O}(n_qn_kh + n_qh)$, where $n_q$ is the number of query nodes and $n_k$ is the number of context nodes.

For \ourzero, the additional computational costs primarily stem from pseudo-context initialization and iterative pseudo-context refinement. The initialization employs K-means clustering on the aligned features, which has a time complexity of $\mathcal{O}(t_{km}nd_un_k)$, where $t_{km}$ is the number of clustering iterations. The iterative process repeats the anomaly scoring and context selection for $T$ rounds. Consequently, the complexity of this phase is $\mathcal{O}(T(n_qn_kh + n_qh + n_q\log n_q))$, where the last term accounts for sorting the query nodes to update context. Since $t_{km}$ and $T$ are small constants, \ourzero maintains a comparable efficiency to \ourmethod.

\section{Algorithm}
\label{app:algo}
The algorithm pseudocode of \ourzero is described in Algo~\ref{alg:arczero_infer_algo}.

\begin{algorithm}[t]
\caption{The Inference Algorithm of \ourzero}
\label{alg:arczero_infer_algo}
\LinesNumbered
\KwIn{Test dataset $\mathcal{D}$ with graph $\mathcal{G}=(\mathcal{V},\mathcal{E},\mathbf{X})$.}
\Param{Well-trained model parameters; propagation iteration $L$; pseudo-context size $n_k$; refinement rounds $T$.}
Align features in $\mathcal{G}$ via the smoothness-based feature alignment\\
Obtain aligned graph $\mathcal{G}=(\mathcal{V},\mathcal{E},\mathbf{X}')$\\
\For{$l = 1:L$}{
    $\mathbf{Z}^{[l]} \gets$ Propagate and transform $\mathbf{X}'=\mathbf{X}^{[0]}$ via Eq.~\eqref{eq:prop_trans}\\
    $\mathbf{R}^{[l]} \gets$ Calculate residual of $\mathbf{Z}^{[l]}$ via Eq.~\eqref{eq:residual}
}
$\mathbf{H} \gets$ Concatenate $[\mathbf{R}^{[1]}|| \cdots|| \mathbf{R}^{[L]}]$ via Eq.~\eqref{eq:residual}\\
Initialize pseudo-context nodes $\mathcal{V}_k^{[1]}$ by running $k$-means on aligned features $\mathbf{X}'$ and selecting $n_k$ pseudo-normal nodes\\
\For{$t = 1:T$}{
    Separate $\mathcal{V}$ into $\mathcal{V}_k^{[t]}$ and $\mathcal{V}_q^{[t]}=\mathcal{V}\setminus \mathcal{V}_k^{[t]}$\\
    Obtain $\mathbf{H}_{k}^{[t]}$ and $\mathbf{H}_{q}^{[t]}$ by indexing $\mathbf{H}$ with $\mathcal{V}_k^{[t]}$ and $\mathcal{V}_q^{[t]}$, respectively\\
    $\tilde{\mathbf{H}}_{q}^{[t]} \gets$ Calculate cross attention from $\mathbf{H}_{q}^{[t]}$ and $\mathbf{H}_{k}^{[t]}$ via Eq.~\eqref{eq:cross_attention}\\
    Compute anomaly scores $f(v_i)^{[t]}$ for all $v_i\in\mathcal{V}_q^{[t]}$ by the distance between $\tilde{\mathbf{H}}_{q}^{[t]}$ and $\mathbf{H}_{q}^{[t]}$\\
    Impute scores of pseudo-context nodes by setting $f(v_i)^{[t]}\gets \min_{v_j\in\mathcal{V}_q^{[t]}}\{f(v_j)^{[t]}\}$ for $v_i\in\mathcal{V}_k^{[t]}$\\
    Update pseudo-context nodes $\mathcal{V}_k^{[t+1]}$ by selecting the top-$k$ (here $k=n_k$) nodes with the smallest scores from $\mathcal{V}_q^{[t]}$
}
Return the final anomaly score $f(v_i)=\frac{1}{T}\sum_{t=1}^{T} f(v_i)^{[t]}$ for each $v_i\in\mathcal{V}$
\end{algorithm}

\section{Additional Experimental Details}
\label{app:exp}
\subsection{Hyperparameter Sensitivity Study}

\begin{figure}[!t]
\vspace{-1.9mm}
\centering
 \subfigure[AUROC]{
   \includegraphics[width=0.45\columnwidth]{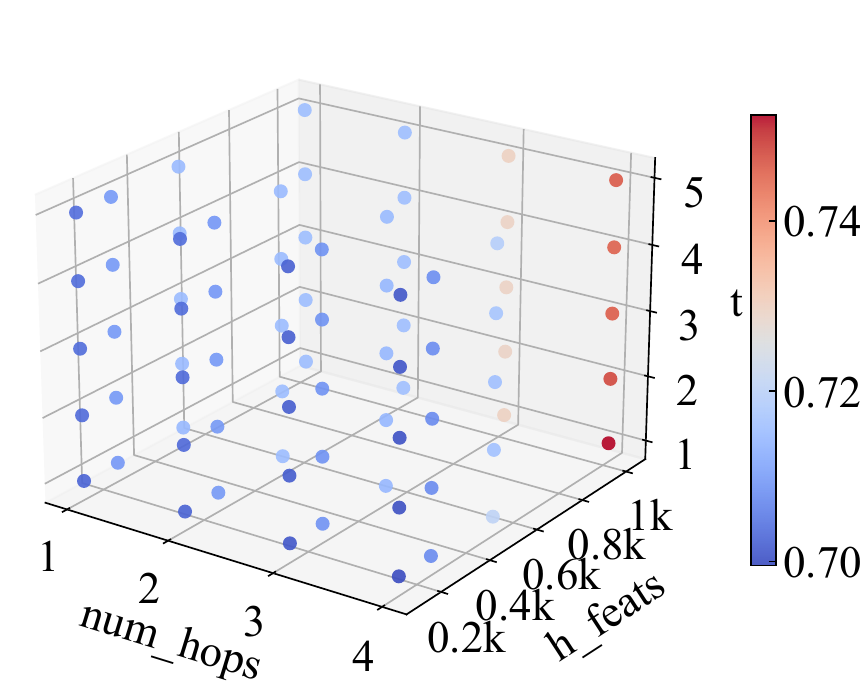}
   \label{subfig:auroc}
}
 \hfill
 \subfigure[AUPRC]{
   \includegraphics[width=0.45\columnwidth]{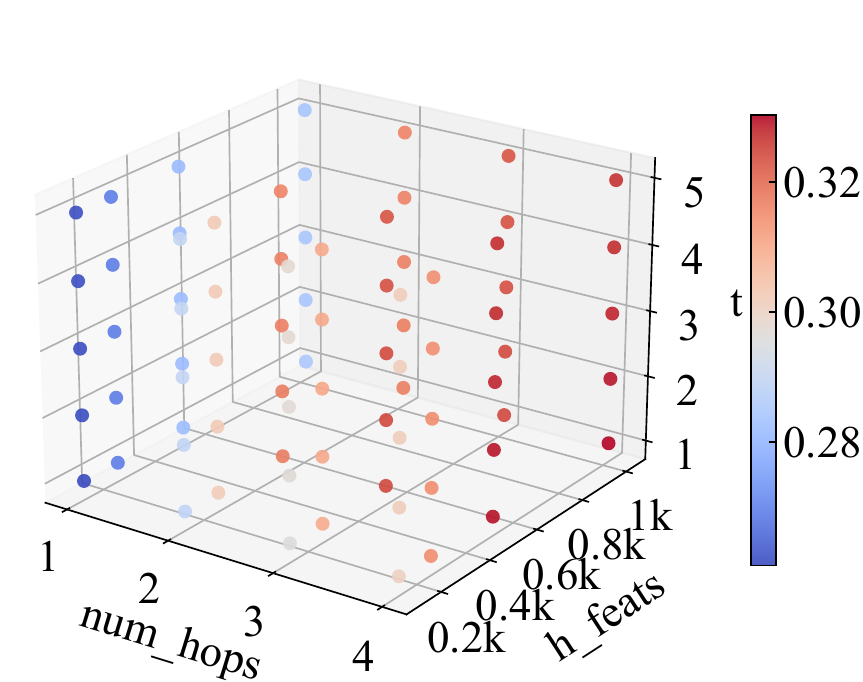}
   \label{subfig:auprc}
}
\caption{Hyperparameter sensitivity analysis.}
\label{fig:hyperparameter_study}\label{fig:hyperparameter_study_appendix}
\end{figure}

The hyperparameter sensitivity results in Fig.~\ref{fig:hyperparameter_study_appendix} indicate that the performance is relatively stable across a broad range of configurations. In general, increasing the representation dimension $h_{\text{feats}}$ tends to improve both AUROC and AUPRC, while varying the propagation depth (\texttt{num\_hops}) and the iteration steps $t$ only introduces mild fluctuations. This suggests that our method does not rely on delicate hyperparameter tuning and remains robust under different settings.

\subsection{AUPRC Performance Comparison}

\begin{table*}[t]
\caption{Anomaly detection performance in terms of AUPRC (in percent, mean$\pm$std). Highlighted are the results ranked \textcolor{firstcolor}{\textbf{\underline{first}}}, \textcolor{secondcolor}{\underline{second}}, and \textcolor{thirdcolor}{\underline{third}}. ``Rank'' indicates the average ranking over 13 datasets. Superscript ``*'' on a baseline method name indicates that our method is significantly different from this baseline on \texttt{ALL\_DATASETS\_TRIAL\_AVG} (paired t-test and Wilcoxon test, p$<$0.05).} 
\vspace{-2mm}
\centering
\label{tab:main_auprc}
\resizebox{1\textwidth}{!}{
\begin{tabular}{l|ccccccc}
\toprule
Method & Cora & CiteSeer & ACM & BlogCatalog & Facebook & Weibo & Reddit \\
\midrule
\rowcolor{Gray}
        \multicolumn{8}{c}{Supervised - Pre-Train Only} \\
GCN & $7.41{\scriptstyle\pm1.55}$ & $6.40{\scriptstyle\pm1.40}$ & $5.27{\scriptstyle\pm1.12}$ & $7.44{\scriptstyle\pm1.07}$ & $1.59{\scriptstyle\pm0.11}$ & $\third{67.21{\scriptstyle\pm15.20}}$ & $3.39{\scriptstyle\pm0.39}$ \\
GAT & $6.49{\scriptstyle\pm0.84}$ & $5.58{\scriptstyle\pm0.62}$ & $4.70{\scriptstyle\pm0.75}$ & $12.81{\scriptstyle\pm2.08}$ & $3.14{\scriptstyle\pm0.37}$ & $33.34{\scriptstyle\pm9.80}$ & $3.73{\scriptstyle\pm0.54}$ \\
BGNN & $4.90{\scriptstyle\pm1.27}$ & $3.91{\scriptstyle\pm1.01}$ & $3.48{\scriptstyle\pm1.33}$ & $5.73{\scriptstyle\pm1.47}$ & $3.81{\scriptstyle\pm2.12}$ & $30.26{\scriptstyle\pm29.98}$ & $3.52{\scriptstyle\pm0.50}$ \\
BWGNN & $7.25{\scriptstyle\pm0.80}$ & $6.35{\scriptstyle\pm0.73}$ & $7.14{\scriptstyle\pm0.20}$ & $8.99{\scriptstyle\pm1.12}$ & $2.54{\scriptstyle\pm0.63}$ & $12.13{\scriptstyle\pm0.71}$ & $3.69{\scriptstyle\pm0.81}$ \\
GHRN & $9.56{\scriptstyle\pm2.40}$ & $7.79{\scriptstyle\pm2.01}$ & $5.61{\scriptstyle\pm0.71}$ & $10.94{\scriptstyle\pm2.56}$ & $2.41{\scriptstyle\pm0.62}$ & $28.53{\scriptstyle\pm7.38}$ & $3.24{\scriptstyle\pm0.33}$ \\
\midrule
\rowcolor{Gray}
        \multicolumn{8}{c}{Unsupervised - Pre-Train Only} \\
DOMINANT & $12.75{\scriptstyle\pm0.71}$ & $13.85{\scriptstyle\pm2.34}$ & $15.59{\scriptstyle\pm2.69}$ & $35.22{\scriptstyle\pm0.87}$ & $2.95{\scriptstyle\pm0.06}$ & $\first{81.47{\scriptstyle\pm0.22}}$ & $3.49{\scriptstyle\pm0.44}$ \\
CoLA & $11.41{\scriptstyle\pm3.51}$ & $8.33{\scriptstyle\pm3.73}$ & $7.31{\scriptstyle\pm1.45}$ & $6.04{\scriptstyle\pm0.56}$ & $1.90{\scriptstyle\pm0.68}$ & $7.59{\scriptstyle\pm3.26}$ & $3.71{\scriptstyle\pm0.67}$ \\
HCM-A & $5.78{\scriptstyle\pm0.76}$ & $4.18{\scriptstyle\pm0.75}$ & $4.01{\scriptstyle\pm0.61}$ & $6.89{\scriptstyle\pm0.34}$ & $2.08{\scriptstyle\pm0.60}$ & $21.91{\scriptstyle\pm11.78}$ & $3.18{\scriptstyle\pm0.23}$ \\
TAM & $11.18{\scriptstyle\pm0.75}$ & $11.55{\scriptstyle\pm0.44}$ & $23.20{\scriptstyle\pm2.36}$ & $10.57{\scriptstyle\pm1.17}$ & $8.40{\scriptstyle\pm0.97}$ & $16.46{\scriptstyle\pm0.09}$ & $3.94{\scriptstyle\pm0.13}$ \\
\midrule
\rowcolor{Gray}
        \multicolumn{8}{c}{Unsupervised - Pre-Train \& Fine-Tune} \\
DOMINANT & $\third{21.35{\scriptstyle\pm0.74}}$ & $\third{23.02{\scriptstyle\pm1.55}}$ & $22.74{\scriptstyle\pm0.95}$ & $\second{35.79{\scriptstyle\pm0.63}}$ & $3.56{\scriptstyle\pm0.15}$ & $\second{77.69{\scriptstyle\pm1.43}}$ & $3.84{\scriptstyle\pm0.74}$ \\
CoLA & $13.91{\scriptstyle\pm5.56}$ & $19.51{\scriptstyle\pm3.73}$ & $8.48{\scriptstyle\pm0.51}$ & $10.43{\scriptstyle\pm1.22}$ & $\second{15.19{\scriptstyle\pm11.04}}$ & $8.03{\scriptstyle\pm1.19}$ & $4.07{\scriptstyle\pm0.13}$ \\
HCM-A & $6.41{\scriptstyle\pm1.33}$ & $4.76{\scriptstyle\pm0.51}$ & $4.41{\scriptstyle\pm0.63}$ & $6.62{\scriptstyle\pm0.14}$ & $2.23{\scriptstyle\pm0.76}$ & $27.20{\scriptstyle\pm5.53}$ & $3.10{\scriptstyle\pm0.19}$ \\
TAM & $13.62{\scriptstyle\pm0.53}$ & $18.66{\scriptstyle\pm1.41}$ & $\first{58.04{\scriptstyle\pm8.17}}$ & $13.90{\scriptstyle\pm0.53}$ & $\third{11.11{\scriptstyle\pm3.20}}$ & $16.47{\scriptstyle\pm0.08}$ & $3.93{\scriptstyle\pm0.09}$ \\

\rowcolor{Gray}
        \multicolumn{8}{c}{Generalist} \\
UNPrompt$^{*}$ & $10.14{\scriptstyle\pm0.97}$ & $11.15{\scriptstyle\pm1.26}$ & $19.36{\scriptstyle\pm2.16}$ & $30.81{\scriptstyle\pm1.46}$ & $8.97{\scriptstyle\pm3.58}$ & $20.27{\scriptstyle\pm1.93}$ & $\third{4.16{\scriptstyle\pm0.19}}$ \\
AnomalyGFM$^{*}$ & $6.40{\scriptstyle\pm0.86}$ & $4.62{\scriptstyle\pm0.31}$ & $4.07{\scriptstyle\pm0.48}$ & $5.21{\scriptstyle\pm0.20}$ & $\first{29.06{\scriptstyle\pm13.56}}$ & $18.48{\scriptstyle\pm5.59}$ & $3.57{\scriptstyle\pm0.33}$ \\
\ourmethod (ours)& $\second{49.33{\scriptstyle\pm1.64}}$ & $\second{45.77{\scriptstyle\pm1.25}}$ & $\second{40.62{\scriptstyle\pm0.10}}$ & $\first{36.06{\scriptstyle\pm0.18}}$ & $8.38{\scriptstyle\pm2.39}$ & $64.18{\scriptstyle\pm0.55}$ & $\first{4.48{\scriptstyle\pm0.28}}$ \\ 
\ourzero (ours)& $\first{49.76{\scriptstyle\pm1.37}}$ & $\first{46.77{\scriptstyle\pm0.84}}$ & $\third{40.12{\scriptstyle\pm0.15}}$ & $\third{35.26{\scriptstyle\pm0.31}}$ & $8.74{\scriptstyle\pm1.86}$ & $64.27{\scriptstyle\pm0.41}$ & $\second{4.34{\scriptstyle\pm0.08}}$ \\
\midrule
\toprule

Method & Amazon & Amazon-Photo & Coauthor-CS & Tolokers & T-Finance & Elliptic & Rank \\
\midrule
\rowcolor{Gray}
        \multicolumn{8}{c}{Supervised - Pre-Train Only} \\
GCN & $6.96{\scriptstyle\pm2.04}$ & $7.50{\scriptstyle\pm0.13}$ & $5.23{\scriptstyle\pm0.14}$ & $\third{26.66{\scriptstyle\pm2.56}}$ & $10.60{\scriptstyle\pm4.13}$ & $2.24{\scriptstyle\pm0.37}$ & $10.1$ \\
GAT & $15.74{\scriptstyle\pm17.85}$ & $8.21{\scriptstyle\pm4.61}$ & $5.51{\scriptstyle\pm1.53}$ & $20.40{\scriptstyle\pm5.03}$ & $4.28{\scriptstyle\pm0.87}$ & $\third{2.73{\scriptstyle\pm0.95}}$ & $9.3$\\
BGNN & $7.51{\scriptstyle\pm0.58}$ & $7.34{\scriptstyle\pm0.59}$ & $5.06{\scriptstyle\pm0.15}$ & $22.25{\scriptstyle\pm4.30}$ & $6.35{\scriptstyle\pm2.27}$ & $2.14{\scriptstyle\pm0.17}$ & $12.1$\\
BWGNN & $13.12{\scriptstyle\pm11.82}$ & $7.55{\scriptstyle\pm1.40}$ & $5.59{\scriptstyle\pm1.84}$ & $20.10{\scriptstyle\pm3.62}$ & $\third{11.67{\scriptstyle\pm8.02}}$ & $2.34{\scriptstyle\pm0.29}$ & $10.0$\\
GHRN & $7.54{\scriptstyle\pm2.01}$ & $7.06{\scriptstyle\pm0.22}$ & $6.65{\scriptstyle\pm0.51}$ & $\first{30.31{\scriptstyle\pm5.24}}$ & $7.72{\scriptstyle\pm4.97}$ & $2.29{\scriptstyle\pm0.37}$ & $9.2$\\
\midrule
\rowcolor{Gray}
        \multicolumn{8}{c}{Unsupervised - Pre-Train Only} \\
DOMINANT & $6.11{\scriptstyle\pm0.29}$ & $11.94{\scriptstyle\pm1.01}$ & $21.30{\scriptstyle\pm0.51}$ & $21.49{\scriptstyle\pm1.47}$ & OOM & OOM & $8.8$\\
CoLA & $11.06{\scriptstyle\pm4.45}$ & $6.74{\scriptstyle\pm0.28}$ & $9.46{\scriptstyle\pm0.32}$ & $23.97{\scriptstyle\pm2.01}$ & $5.04{\scriptstyle\pm0.75}$ & OOM & $10.8$\\
HCM-A & $5.87{\scriptstyle\pm0.07}$ & $5.90{\scriptstyle\pm0.10}$ & $2.63{\scriptstyle\pm0.09}$ & $22.64{\scriptstyle\pm0.17}$ & $7.19{\scriptstyle\pm0.94}$ & OOM & $14.0$\\
TAM & $10.75{\scriptstyle\pm3.10}$ & $8.56{\scriptstyle\pm1.90}$ & $17.99{\scriptstyle\pm2.94}$ & $22.84{\scriptstyle\pm0.06}$ & OOM & OOM & $8.7$\\
\midrule
\rowcolor{Gray}
        \multicolumn{8}{c}{Unsupervised - Pre-Train \& Fine-Tune} \\
DOMINANT & $7.48{\scriptstyle\pm0.46}$ & $\third{22.18{\scriptstyle\pm0.29}}$ & $23.30{\scriptstyle\pm0.22}$ & $20.38{\scriptstyle\pm1.99}$ & OOM & OOM & $\third{7.2}$\\
CoLA & $7.27{\scriptstyle\pm1.13}$ & $9.21{\scriptstyle\pm0.31}$ & $\third{25.77{\scriptstyle\pm3.11}}$ & $25.29{\scriptstyle\pm3.02}$ & $2.91{\scriptstyle\pm0.04}$ & OOM & $7.6$\\
HCM-A & $5.64{\scriptstyle\pm0.09}$ & $6.04{\scriptstyle\pm0.25}$ & $2.88{\scriptstyle\pm0.27}$ & $21.85{\scriptstyle\pm2.14}$ & OOM & OOM & $14.2$\\
TAM & $11.56{\scriptstyle\pm1.80}$ & $7.31{\scriptstyle\pm0.63}$ & $18.01{\scriptstyle\pm2.93}$ & $22.97{\scriptstyle\pm0.07}$ & OOM & OOM & $7.7$\\

\rowcolor{Gray}
        \multicolumn{8}{c}{Generalist} \\
UNPrompt$^{*}$ & $9.57{\scriptstyle\pm1.74}$ & $18.11{\scriptstyle\pm3.96}$ & $13.94{\scriptstyle\pm2.99}$ & $19.37{\scriptstyle\pm0.52}$ & $2.64{\scriptstyle\pm0.07}$ & $2.22{\scriptstyle\pm0.10}$ & $7.9$\\
AnomalyGFM$^{*}$ & $\first{54.70{\scriptstyle\pm1.11}}$ & $6.80{\scriptstyle\pm0.25}$ & $2.83{\scriptstyle\pm0.20}$ & $\second{26.82{\scriptstyle\pm1.71}}$ & $11.18{\scriptstyle\pm2.84}$ & $1.74{\scriptstyle\pm0.19}$ & $10.2$\\
\ourmethod (ours)& $\third{44.25{\scriptstyle\pm7.41}}$ & $\first{33.02{\scriptstyle\pm1.33}}$ & $\first{36.51{\scriptstyle\pm0.12}}$ & $23.53{\scriptstyle\pm0.79}$ & $\second{19.00{\scriptstyle\pm4.44}}$ & $\first{8.47{\scriptstyle\pm0.51}}$ & $\first{2.6}$\\ 
\ourzero (ours)& $\second{49.21{\scriptstyle\pm2.29}}$ & $\second{29.69{\scriptstyle\pm5.70}}$ & $\second{36.34{\scriptstyle\pm0.35}}$ & $23.05{\scriptstyle\pm0.32}$ & $\first{24.52{\scriptstyle\pm0.70}}$ & $\second{8.42{\scriptstyle\pm0.1}}$ & $\second{2.7}$\\
\bottomrule
\end{tabular}
}
\vspace{-2mm}
\end{table*}

Table~\ref{tab:main_auprc} reports the AUPRC results and leads to conclusions consistent with the AUROC comparison in the main text. Overall, \ourmethod and \ourzero achieve the strongest AUPRC performance on most datasets and obtain top average ranks, demonstrating robust generalization under both few-shot and zero-shot settings. In particular, our methods outperform the latest generalist baselines (UNPrompt and AnomalyGFM) in overall AUPRC across the test suite.

\subsection{Additional Shot-Number Analysis}

\begin{figure}[!t]
 \centering
 \subfigure[CiteSeer]{
   \includegraphics[width=0.45\columnwidth]{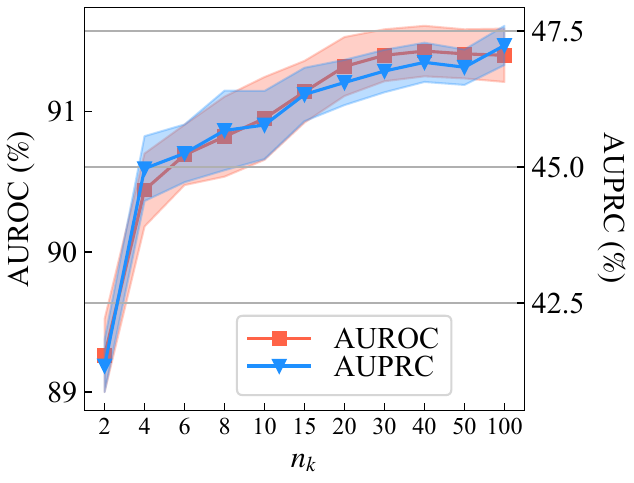}
 }
 \hfill
 \subfigure[Amazon]{
   \includegraphics[width=0.45\columnwidth]{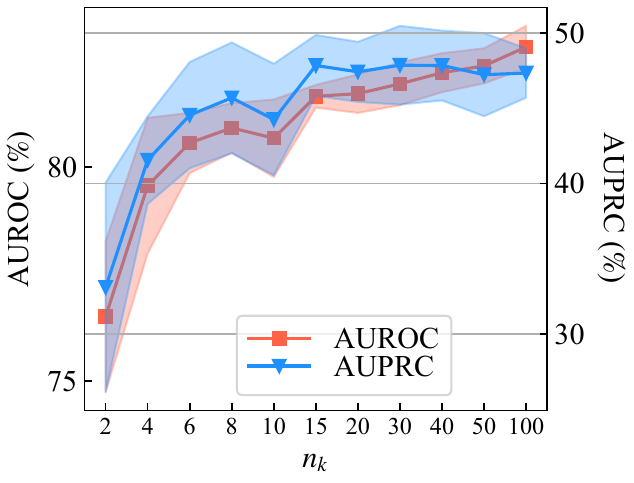}
 }
 \caption{Additional results for varying $n_k$ in the few-shot setting (\ourmethod).}
 \vspace{-2mm}
 \label{fig:shot_num_fs_appendix}
\end{figure}

These additional few-shot results are consistent with the main observations: increasing $n_k$ generally improves performance and then saturates. In particular, CiteSeer shows a clearer monotonic gain with larger $n_k$, while Amazon exhibits a faster saturation and slightly larger fluctuations, suggesting that a moderate number of context nodes is sufficient in practice.

\begin{figure}[!t]
 \centering
 \subfigure[CiteSeer]{
   \includegraphics[width=0.45\columnwidth]{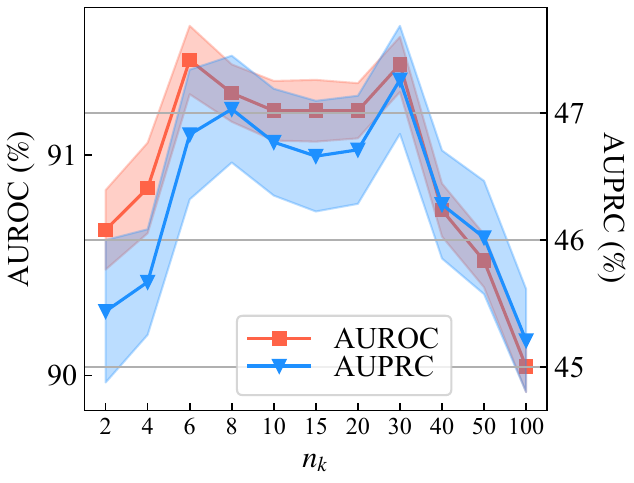}
 }
 \hfill
 \subfigure[Amazon]{
   \includegraphics[width=0.45\columnwidth]{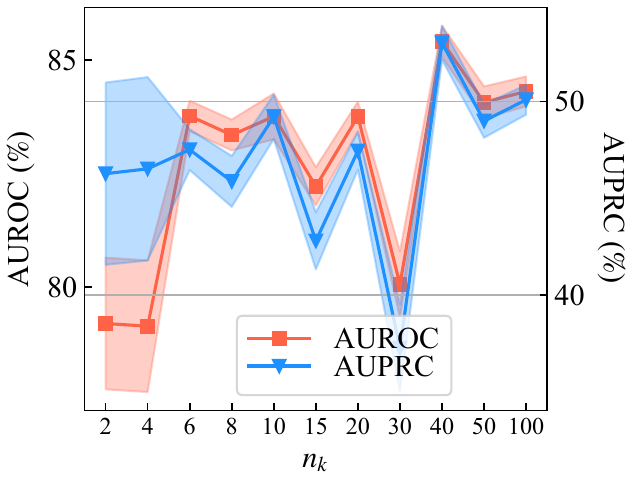}
 }
 \vspace{-2mm}
 \caption{Additional results for varying $n_k$ in the zero-shot setting (\ourzero).}
 \label{fig:shot_num_zs_appendix}
\end{figure}

Compared with the few-shot case, the zero-shot curves are more sensitive to $n_k$, since pseudo-context quality depends on the selected pseudo-normal nodes. We observe that CiteSeer achieves stable performance within a moderate range of $n_k$, while Amazon shows more noticeable fluctuations when $n_k$ becomes large, indicating that overly large pseudo-context sets may introduce noisy nodes and reduce stability.




\vfill

\end{document}